\documentclass[conf]{new-aiaa}

\usepackage[utf8]{inputenc} 
\usepackage[T1]{fontenc}
\usepackage{textcomp}
\usepackage[version=4]{mhchem}
\usepackage{siunitx}

\usepackage{lineno,hyperref}
\usepackage{bm}
\usepackage{amsmath, amsfonts}
\usepackage{longtable, tabularx, booktabs}
\usepackage{makecell}
\usepackage{multirow}
\usepackage{xcolor}
\usepackage{graphicx}
\graphicspath{ {./figures/} }
\usepackage{caption}
\usepackage{subcaption}

\newcommand{\synthetic}{\texttt{synthetic} }
\newcommand{\lightbox}{\texttt{lightbox} }
\newcommand{\sunlamp}{\texttt{sunlamp} }

\newcommand{\Ehead}{h_\text{E}}
\newcommand{\Hhead}{h_\text{H}}
\newcommand{\Shead}{h_\text{S}}

\newcommand{\roe}{\delta\bm{\alpha}}

\newcommand{\mrp}{\delta\bm{p}}
\newcommand{\dq}{\delta\bm{q}}

\usepackage{soul}

\usepackage[most]{tcolorbox}
\newtcolorbox{highlighted}{colback=yellow,breakable,boxrule=0pt,frame empty,
    boxsep=0pt,left=0pt,right=0pt,top=0pt,bottom=0pt}

\title{Adaptive Neural Network-based Unscented Kalman Filter for Robust Pose Tracking of Noncooperative Spacecraft}

\author{Tae Ha Park\footnote{Ph.D.~Candidate, Department of Aeronautics \& Astronautics, 496 Lomita Mall; tpark94@stanford.edu. Student Member AIAA.} and Simone D'Amico\footnote{Associate Professor, Department of Aeronautics \& Astronautics, 496 Lomita Mall. Associate Fellow AIAA.}}
\affil{Stanford University, Stanford, CA 94305}

\begin{document}

\maketitle

\begin{abstract}
This paper\footnote[3]{Presented as Paper AAS 22-554 at the 2022 AAS/AIAA Astrodynamics Specialist Conference, Charlotte, North Carolina, August 7-11.} presents a neural network-based Unscented Kalman Filter (UKF) to estimate and track the pose (i.e., position and orientation) of a known, noncooperative, tumbling target spacecraft in a close-proximity rendezvous scenario. The UKF estimates the target's orbit and attitude relative to the servicer based on the pose information provided by a multi-task Convolutional Neural Network (CNN) from incoming monocular images of the target. In order to enable reliable tracking, the process noise covariance matrix of the UKF is tuned online using adaptive state noise compensation which leverages a newly developed closed-form process noise model for relative attitude dynamics. This paper also introduces the Satellite Hardware-In-the-loop Rendezvous Trajectories (SHIRT) dataset to enable comprehensive analyses of the performance and robustness of the proposed pipeline. SHIRT comprises the labeled images of two representative rendezvous trajectories in low Earth orbit created using both a graphics renderer and a robotic testbed. Specifically, the CNN is solely trained on synthetic data, whereas functionality and performance of the complete navigation pipeline are evaluated on real images from the robotic testbed. The proposed UKF is evaluated on SHIRT and is shown to have sub-decimeter-level position and degree-level orientation errors at steady-state.
\end{abstract}

\section*{Nomenclature}

{\renewcommand\arraystretch{1.0}
	\noindent\begin{longtable*}{@{}l @{\quad=\quad} l@{}}
        $\bullet^\text{diag}$, $\text{diag}(\bullet)$ & vector of the diagonal elements of a matrix \\
        $\bm{K}$ & Kalman gain matrix \\
        $N_W$ & length of the sliding window in covariance matching \\
        $\bm{t}_{B/A}^C$ & translation vector from the origin of reference frame $A$ to that of $B$ expressed in $C$ \\
        $\bm{P}$ & Kalman filter state covariance matrix \\
        $\bm{Q}$ & process noise covariance matrix \\
        $\widetilde{\bm{Q}}$ & process noise power spectral density matrix \\
        $\bm{q}_{B/A}$ & quaternion vector aligning the reference frame $A$ to $B$ \\
        $\bm{R}_{B/A}$ & direction cosine matrix corresponding to $\bm{q}_{B/A}$ \\
        $\bm{S}$ & Kalman filter measurement covariance matrix \\
        $t$ & time \\
        $\bullet^\text{vech}$, $\text{vech}(\bullet)$ & vector of the lower-triangular elements of a symmetric matrix \\
        $\bm{x}$ & Kalman filter state vector \\
        $\bm{y}$ & Kalman filter measurement vector \\
        $\bm{\alpha}$ & orbital elements \\
        $\bm{\Gamma}$ & process noise mapping matrix \\
        $\bm{\Delta}^x$ & Kalman filter state correction term \\
        $\bm{\Delta}^y$ & Kalman filter pre-fit residual or measurement innovation \\
        $\bm{\roe}$ & relative orbital elements \\
        $\dq$ & error-quaternion vector \\
        $\mrp$, $\mrp^S$ & original and shadow modified rodrigues parameters \\
        $\bm{\omega}_{B/A}^C$ & relative angular velocity of the reference frame $B$ relative to $A$ expressed in $C$ \\
        $\bm{\Phi}$ & state transition matrix \\
		\multicolumn{2}{@{}l}{Subscripts}\\
        $C$ & camera reference frame \\
        $\text{E}$ & pertaining to the vector pose measurements from the EfficientPose head of SPNv2 \\
        $\text{H}$ & pertaining to the heatmap measurements from the Heatmap head of SPNv2 \\
        $k$ & Kalman filter discrete time step \\
        $S$ & servicer reference frame centered at its center-of-mass and aligned with its principal axes \\
        $T$ & target reference frame centered at its center-of-mass and aligned with its principal axes \\
\end{longtable*}}

\section{Introduction}

\lettrine{T}{he} on-board estimation and tracking of the pose (i.e., position and orientation) of a target Resident Space Object (RSO) is a key enabling technology for various on-orbit servicing \citep{Reed2016RestoreL} and active debris removal \citep{Forshaw2016Removedebris} missions. In these missions, real-time information about the target's pose with respect to the servicer spacecraft is required in order to plan and execute safe, autonomous, and fuel-efficient rendezvous and docking trajectories. Extracting pose from a single or sequence of images captured with a low Size-Weight-Power-Cost (SWaP-C) sensor such as a monocular camera is especially attractive in comparison to more complex sensor systems such as Light Detection and Ranging (LiDAR) or stereovision. This paper considers the case of monocular pose tracking of a single known, noncooperative, possibly tumbling target satellite, which is representative of servicing and lifetime extension missions.

Existing approaches to spacecraft pose estimation using monocular images rely on Machine Learning (ML) and particularly Convolutional Neural Networks (CNN) to learn the implicit mapping between an image and its pose information \citep{Sharma2020TAES, Park2019AAS, PasqualettoCassinis2021Coupled, Black2021PoseEstimationCygnus, Chen2019SatellitePE, Proenca2019Photorealistic}. However, these approaches require training datasets which consist of a large number of labeled images of the target RSO in various operational conditions and environments. This is prohibitively expensive for the spaceborne applications. Therefore, the available datasets for training spaceborne ML models depend almost exclusively on computer graphics engines to render synthetic images in large amounts. Key examples include OpenGL for images of the PRISMA mission's Tango spacecraft \citep{PRISMA_chapter} in the SPEED dataset \citep{Sharma2020TAES, Kisantal2020SPEC, Sharma2019SPEEDonSDR}, Cinema 4D for the Envisat spacecraft \citep{PasqualettoCassinis2020CNNEKF, PasqualettoCassinis2022ORGL}, Blender for the Cygnus spacecraft \citep{Black2021PoseEstimationCygnus}, and Unreal Engine for Soyuz in the URSO dataset \citep{Proenca2019Photorealistic}. However, as evidenced by the result of the first Satellite Pose Estimation Competition (SPEC2019) \citep{Kisantal2020SPEC} organized by the authors, the models trained exclusively on synthetic images suffer from the \emph{domain gap} problem \citep{BenDavid2010LearningFromDiffDomains, Peng2017VisDA} as neural networks may overfit features which are specific to the synthetic imagery used for training. Domain gap causes severely degraded performance when operating on image domains with a dissimilar data distribution (e.g., real spaceborne images) compared to that of the training dataset. Moreover, even if a CNN can be trained to be robust across domain gaps (as showcased in literature \cite{Park2021speedplus, Black2021PoseEstimationCygnus}), there remains the question of validation: without access to space-based testing, how can one demonstrate that a CNN is truly robust to \emph{spaceborne} images of the target prior to deployment?

This important question was recently addressed experimentally with the authors' SPEED+ dataset, which made it possible to comprehensively analyze a CNN's robustness across the domain gap for spaceborne navigation \citep{Park2021speedplus, Park2021speedplusSDR}. In addition to 60,000 synthetic images, SPEED+ also includes nearly 10,000 images of a Tango mockup model captured using the robotic Testbed for Rendezvous and Optical Navigation (TRON) facility at Stanford's Space Rendezvous Laboratory (SLAB). These Hardware-In-the-Loop (HIL) images and accompanying high-accuracy pose labels are captured within a high-fidelity space simulation facility equipped with calibrated light boxes and a metal halide sun lamp that respectively simulate diffuse and direct light as typically encountered in orbit. The baseline study by \citet{Park2021speedplus} showed that the HIL images can be used as on-ground surrogates of otherwise unavailable spaceborne images. The SPEED+ dataset was used for the second Satellite Pose Estimation Competition (SPEC2021) \citep{park2023spec2021} with emphasis on bridging the domain gap between the synthetic training and HIL test images. Specifically, the pose labels of the HIL images were kept private, so that participants were forced to design robust pose estimation algorithms using only the labeled synthetic images and optionally the unlabeled HIL images as one would during the preliminary phases of a space mission.

In response to the domain gap challenge posed by SPEED+, \citet{park2022spnv2} recently proposed the Spacecraft Pose Network v2 (SPNv2) to bridge the domain gap in SPEED+ via a multi-task learning architecture and extensive data augmentation. SPNv2 consists of a shared, multi-scale feature encoder followed by multiple prediction heads that each perform different yet related tasks such as bounding box prediction, pose regression,satellite foreground segmentation, and heatmap prediction around surface keypoints defined in advance. SPNv2 is trained exclusively on synthetic images and is shown to generalize better to unseen HIL images when it is jointly trained on different ML tasks with exclusive image data augmentation, such as random solar flare and style augmentation \citep{Jackson2019ICCV_StyleAug}. 


The capability of SPNv2 has so far only been showcased on a single-image basis. In fact, very few approaches for monocular spacecraft rendezvous extend the application of CNN beyond single images to trajectories and video streams. Some examples include \citet{Proenca2019Photorealistic} who qualitatively test their CNN on a video of the Soyuz spacecraft captured in LEO and \citet{PasqualettoCassinis2022ORGL} who test their CNN on 100 images of the Envisat mockup spacecraft captured at the Orbital Robotics and GNC laboratory of ESTEC. \citet{PasqualettoCassinis2021Coupled} integrate a CNN into an Extended Kalman Filter (EKF) but only perform testing on a trajectory of synthetic images. To the authors' knowledge, there is currently no work that simultaneously achieves: (1) integration of CNN or any ML models into a navigation filter for space missions; and (2) quantitative evaluation of performance and robustness of the integrated system when operating on spacecraft trajectory images which originate from a different source to the exclusively synthetic training images. The goal of this paper is to overcome the two aforementioned challenges by leveraging and building upon the unique tools, assets and models developed by the authors such as TRON, SPEED+ and SPNv2.

The primary contribution of this paper is the integration of SPNv2 into an Unscented Kalman Filter (UKF) \citep{Julier2004UnscentedFiltering} to enable continuous, stable pose tracking of a noncooperative spacecraft from a sequence of monocular images during a rendezvous phase. The proposed UKF tracks the pose of the target spacecraft relative to the servicer, which consists of the target's 6D orbital state, orientation, and angular velocity. In order to reliably track relative orientation within the Kalman filter framework, a technique from the Multiplicative Extended Kalman Filter (MEKF) \citep{Tweddle2015MEKF, Sharma2017AASGCC} and Unscented Quaternion Estimator (USQUE) \citep{Crassidis2003UnscentedAttitude} is adopted, in which the UKF specifically estimates the Modified Rodrigues Parameter (MRP) \citep{Schaub1995GRP} associated with the error-quaternion of the relative orientation between subsequent time updates. To further stabilize filter convergence amidst time-varying noise due to target tumbling and consequently time-varying illumination conditions), the process noise covariance matrix $\bm{Q}$ is adjusted at each iteration using Adaptive State Noise Compensation (ASNC) \citep{Stacey2021ASNC}. ASNC is a novel technique developed by \citet{Stacey2021ASNC} that solves for an optimal positive semi-definite $\bm{Q}$ based on estimates from Covariance Matching (CM) \citep{Myers1976CovarianceMatching} and the underlying continuous-time dynamics. Specifically, in addition to the process noise covariance matrix models derived for various orbital states by \citet{Stacey2022CovModel}, an analytical model for the process noise of relative attitude dynamics is newly derived and implemented in this paper.

The secondary contribution of this paper is the Satellite Hardware-in-the-loop Rendezvous Trajectories (SHIRT) dataset \cite{park2022shirt}. The SHIRT dataset\footnote{\url{https://taehajeffpark.com/shirt/}.} consists of two rendezvous trajectory scenarios (denoted ROE1 and ROE2) in Low Earth Orbit (LEO) created from two different image sources. The first is the OpenGL-based computer graphics renderer used to synthesize the \texttt{synthetic} dataset of SPEED+. The other is the TRON facility illuminated with Earth albedo light boxes as used to create the \texttt{lightbox} domain imagery of SPEED+. In ROE1, the servicer maintains an along-track separation typical of a standard v-bar hold point while the target rotates about one principal axis. In ROE2, the servicer slowly approaches the target which tumbles about two principal axes. The SHIRT dataset is employed to evaluate the performance of the SPNv2-integrated UKF across the domain gap. It is shown that the UKF with ASNC and SPNv2 trained on the SPEED+ $\synthetic$ training set is able to achieve sub-decimeter-level position and degree-level orientation error at steady-state on the SHIRT $\lightbox$ images which SPNv2 has not seen during its training phase. Extensive Monte Carlo simulations and analyses show that the filter is robust across the domain gap despite imperfect absolute state knowledge of the servicer and also conforms to the docking requirements presented by a previous mission example. To the authors' knowledge, this is the first time a CNN's performance across the domain gap has been systematically tested on spacecraft trajectory images whilst integrated into a navigation filter.

This paper is outlined as follows. First, Section \ref{sec:Development Pipeline} describes the newly proposed navigation pipeline which combines a navigation filter with an ML module for monocular proximity operations in space. Then, Section \ref{sec:Preliminaries} provides preliminaries for the SPNv2 model and the adopted UKF formulations to aid understanding of the subsequent algorithmic contributions. These are presented in Sec.~\ref{sec:Process Noise Modeling}, including the analytical process noise models for relative orbital and attitude motion. The detailed characteristics, simulation parameters and image acquisition processes of the novel SHIRT dataset are outlined in Sec.~\ref{sec:SHIRT}. Section \ref{section:experiments} analyzes the performance and robustness of the proposed UKF with SPNv2 on the HIL trajectory images of SHIRT. The paper ends with conclusions in Sec.~\ref{sec:conclusion}.
\section{Complete Navigation Pipeline} \label{sec:Development Pipeline}

\begin{figure}[!t]
	\centering
	\includegraphics[width=1.0\textwidth]{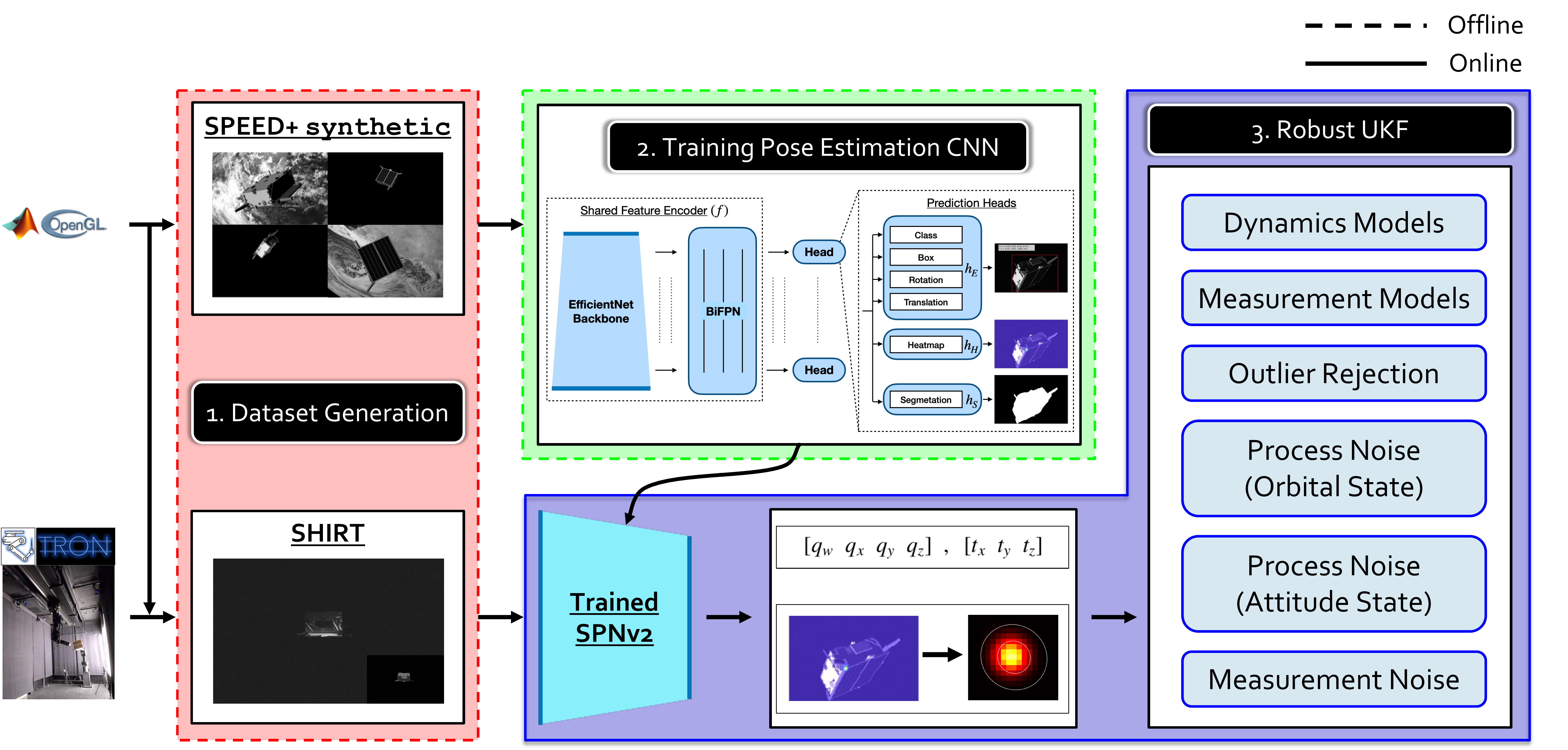}
	\caption{The proposed strategy for development and on-ground validation of a UKF with a CNN-based image processing module. The strategy consists of three stages: 1) dataset generation; 2) CNN training; and 3) robust UKF.}
	\label{fig:architecture}
\end{figure}

This section presents the complete navigation pipeline designed to integrate machine learning and nonlinear estimation algorithms for spaceborne proximity rendezvous missions. The overarching strategy is visualized in Fig.~\ref{fig:architecture} and consists of three steps: 1) dataset generation; 2) CNN training; and 3) CNN inference and a robust UKF. Operationally, the first two steps are completed offline or on-ground in the context of space missions, and the last step is performed online or in orbit. Note that while Fig.~\ref{fig:architecture} describes specific tools and models, the pipeline is readily generalizable to any other renderer, target model, CNN architecture or Kalman filter framework. In this work, it is assumed that navigation is performed about a known spaceborne target whose 3D model is available.

The first step of this pipeline is to generate datasets that can be used to 1) train and validate a CNN for monocular spacecraft pose estimation and 2) test the performance and robustness of a navigation filter using the trained CNN as its image processing module. These datasets must contain images of the target from two distinct sources. One is synthetic imagery created with a computer renderer and the target's CAD or mesh model for the purpose of training neural networks and validating their functionality. The other is HIL imagery which exhibits high-fidelity spaceborne-like visual characteristics in order to test the robustness of neural networks across domain gaps. The HIL images are created with a robotic testbed such as TRON \citep{Park2021AAS} and are intended to replace otherwise unavailable spaceborne images. To train the CNN, one can use a public benchmark dataset such as the authors' SPEED+ \citep{Park2021speedplus, Park2021speedplusSDR} which contains both imagery and pose labels. Its $\synthetic$ domain contains 60,000 images of the PRISMA mission's Tango spacecraft \citep{PRISMA_chapter} rendered with OpenGL and is used to train the CNN. Its two other HIL domains -- $\lightbox$ and $\sunlamp$ -- are created with TRON and contain nearly 10,000 labeled images that can be used to evaluate a CNN's robustness. To further test a navigation filter with the CNN, one must construct a dataset of sequential images from representative rendezvous scenarios. As such dataset is not available in the literature, this work introduces a novel dataset called Satellite Hardware-In-the-loop Rendezvous Trajectories (SHIRT) whose full details are provided in Sec.~\ref{sec:SHIRT}.



After rigorous dataset generation, the next step is to design and train a CNN for spacecraft pose estimation that is robust across domain gaps. This work uses SPNv2 \citep{park2022spnv2} developed by the authors, as it is designed specifically to address the domain gap challenge. SPNv2 is trained exclusively on the SPEED+ $\synthetic$ training set with extensive data augmentation and is shown to have good performance on the SPEED+ HIL domain images without having seen them during the training phase. As noted in Fig.~\ref{fig:architecture}, training is done offline on the ground with dedicated computing power such as Graphics Processing Units (GPU). The CNN is not intended to be trained as a whole onboard the satellite due to the limited computing power of satellite avionics. It may be possible to re-train a very small subset of the neural network's parameters, but such a scenario is not considered in this work. More details on SPNv2 can be found in Sec.~\ref{subsec:SPNv2}.

The final step is to design a robust navigation filter and incorporate the trained CNN as an image processing module. This work employs a UKF to estimate the relative orbit and attitude of the Tango spacecraft via the pose-related measurements extracted by SPNv2. The performance of the UKF is evaluated on the HIL trajectory images of SHIRT. In order to be robust towards any outlier measurements provided by SPNv2 due to domain gap effects, the UKF employs a number of innovative features. These include an outlier rejection scheme based on the squared Mahalanobis distance criterion \citep{Tweddle2015MEKF}, adaptive tuning of the process noise covariance matrix via ASNC \citep{Stacey2021ASNC} for both relative orbit and attitude states, and online estimation of the measurement noise associated with predicted spacecraft model keypoints based on the shapes of heatmaps extracted by SPNv2 \citep{PasqualettoCassinis2021Coupled}. ASNC for relative attitude motion is enabled by deriving a new analytical model for its associated process noise (see Sec.~\ref{sec:Process Noise Modeling}). 
\section{Preliminaries} \label{sec:Preliminaries}

This section provides a preliminary description of the SPNv2 model and components of the proposed UKF framework. It also provides the background of ASNC to aid understanding of the derivation of new analytical process noise models in the next section.

\subsection{SPNv2} \label{subsec:SPNv2}

\begin{figure}[!t]
	\centering
	\includegraphics[width=0.45\textwidth]{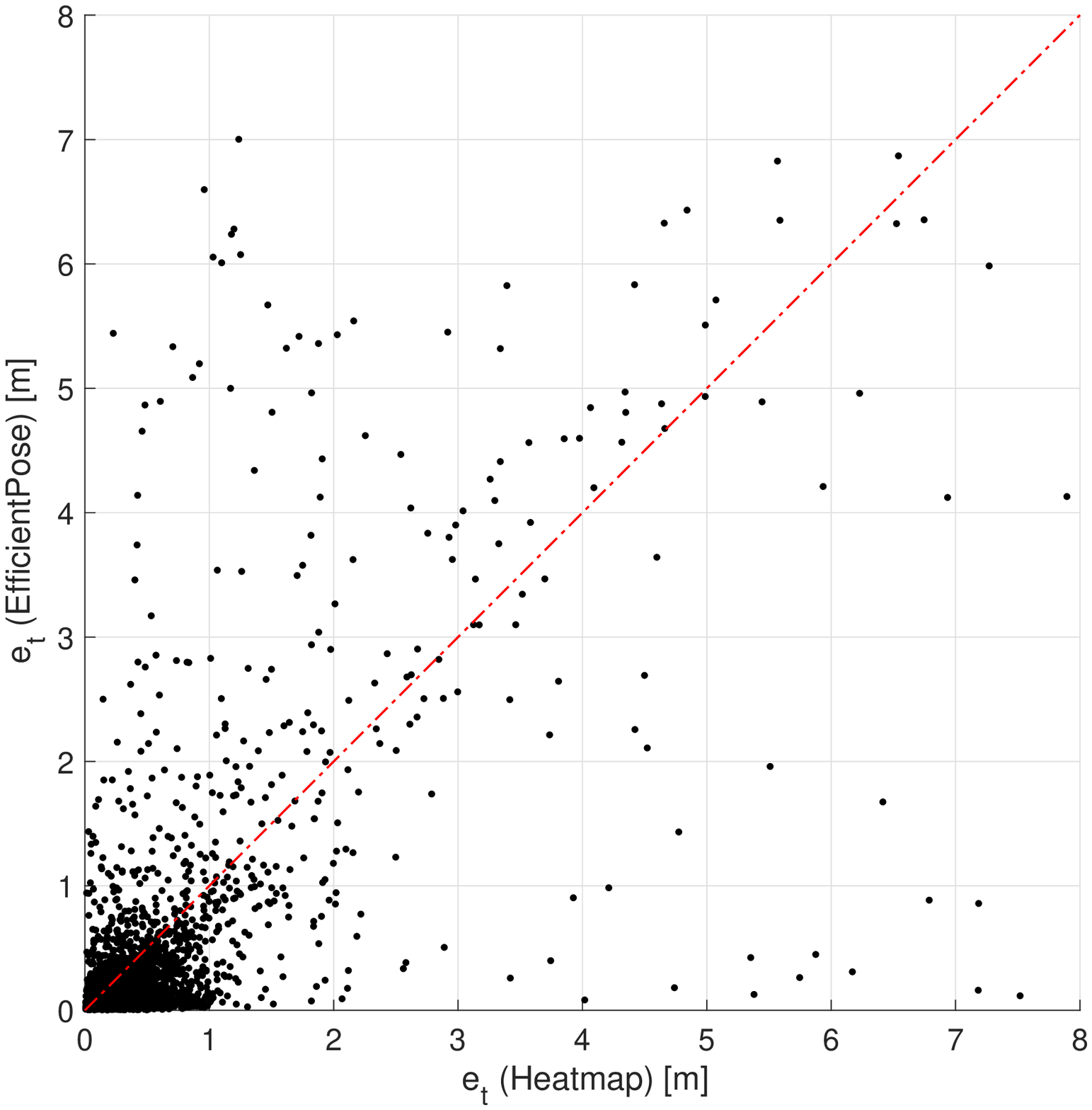}
	\includegraphics[width=0.468\textwidth]{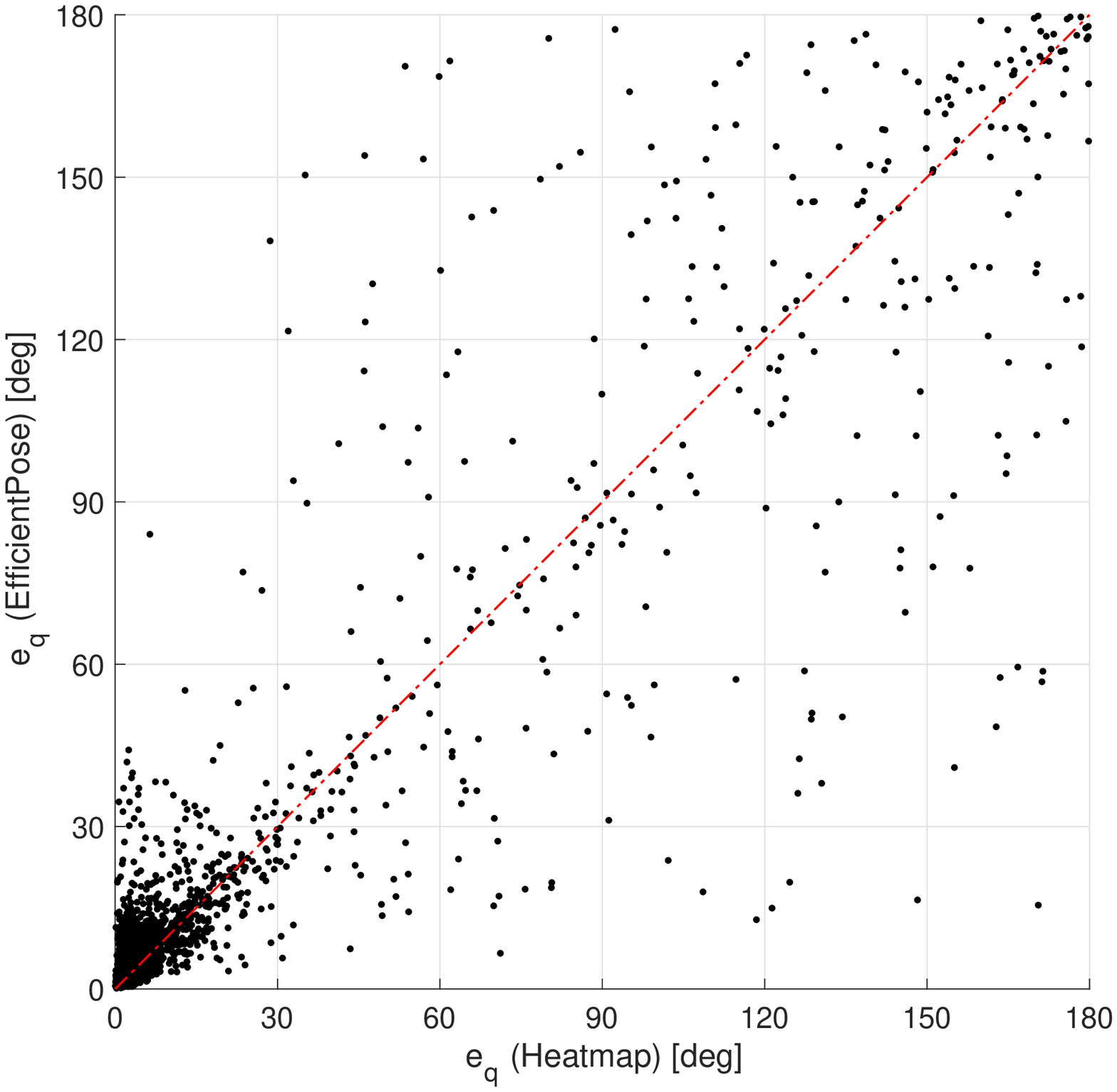}
	\caption{The comparison of $e_\text{t}$ and $e_\text{q}$ for translation and rotation predicted from $h_\text{H}$ and $h_\text{E}$, respectively, on the SPEED+ $\lightbox$ domain. The red line indicates the identity function.}
	\label{fig:heatmap_efficientpose_comparison}
\end{figure}

As visualized in the second stage of Fig.~\ref{fig:architecture}, SPNv2 \citep{park2022spnv2} is a multi-scale, multi-task learning CNN with a shared feature encoder based on the EfficientNet \citep{Tan2019EfficientNetICML} backbone and BiFPN layers \citep{Tan2020EfficientDetCVPR} which fuse features at different scales. The output of the feature encoder is provided to multiple prediction heads which perform different tasks that are not necessarily related to pose estimation. Namely, the EfficientPose head ($\Ehead$) follows the implementation of EfficientPose \citep{Bukschat2020EfficientPose} to predict the bounding box around the spacecraft and directly regress the translation and orientation vectors of the target. The Heatmap head ($\Hhead$) outputs $K$ heatmaps of size $H \times W$ whose peaks are associated with the 2D projected locations of $K$ pre-designated keypoints of the target spacecraft. Finally, the Segmentation head ($\Shead$) performs binary pixel-wise classification of the spacecraft foreground. All prediction heads and the feature encoder of SPNv2 are jointly trained on the SPEED+ $\synthetic$ training set during the offline training phase with extensive data augmentation including random solar flare and style augmentation \citep{Jackson2019ICCV_StyleAug}. \citet{park2022spnv2} shows through extensive ablation studies that SPNv2 trained exclusively on SPEED+ $\synthetic$ images achieves low pose errors on both HIL image sets of SPEED+ and that it owes its success to the multi-task learning architecture and data augmentation.

Given the unique multi-task learning structure of SPNv2, the pose predictions can be retrieved from the outputs of either $\Ehead$ or $\Hhead$. Specifically, given the known 3D coordinates of $K$ keypoints in the target model's reference frame, the corresponding 2D keypoint locations can be extracted from the peaks of heatmaps from $\Hhead$, which are then converted to 6D pose by solving Perspective-$n$-Point (P$n$P) \cite{Sharma2016CompAssessment} along with their corresponding 3D coordinates. Recall that $\Ehead$ provides the pose directly. To compare the poses retrieved from both prediction heads across the domain gap, Figure \ref{fig:heatmap_efficientpose_comparison} evaluates the translation error ($e_\text{t}$) and rotation error ($e_\text{q}$) of the outputs from $\Hhead$ via EP$n$P \cite{Lepetit2008EPnP} and $\Ehead$, respectively, by SPNv2 on the SPEED+ $\lightbox$ test set. The errors for individual samples are defined as 
\begin{subequations} \label{eqn:pos rot metric individual}
	\begin{align} 
		e_\text{t} &= \| \bm{\hat{t}} - \bm{t} \| \\
		e_\text{q} &= 2 \arccos \big( | \langle \hat{\bm{q}}, \bm{q} \rangle | \big)
	\end{align}
\end{subequations}
where $(\bm{\hat{t}}, \bm{\hat{q}})$ and $(\bm{t}, \bm{q})$ are respectively the predicted and ground-truth translation and quaternion vectors. It is evident from Fig.~\ref{fig:heatmap_efficientpose_comparison} that for most samples, predictions made from one head are often better than the other with a weak correlation of the errors. Therefore, the redundant pose information from one SPNv2 prediction head can be used to hedge against the failure of another. To that end, outputs of both $\Ehead$ and $\Hhead$ are provided as independent measurements to the navigation filter in this work. This work uses a batch-agnostic variant of the SPNv2 architecture which encompasses approximately 52.5M learnable parameters in the feature encoder and is built with Group Normalization (GN) layers \citep{Wu2018ECCV_GroupNorm} throughout the network. For more information on SPNv2, its characteristics, and its training procedure, readers are referred to \citet{park2022spnv2}.  


\subsection{UKF Dynamics Model}

In order to reliably estimate and update the target's orientation, techniques from the MEKF \cite{Tweddle2015MEKF, Sharma2017AASGCC, Beierle2018TwoStageAttitude} and USQUE \cite{Crassidis2003UnscentedAttitude} are adopted in which the UKF state vector includes the Modified Rodrigues Parameter (MRP) \cite{Schaub1995GRP} associated with the error-quaternion of the relative orientation between subsequent time updates. The UKF state vector describes the relative motion of the target (subscript $T$) with respect to the servicer (subscript $S$) and is given as
\begin{equation} \label{eqn:state_vector}
	\bm{x} = \left[ ~~ \delta\bm{\alpha}^\top ~~ \delta\bm{p}^\top ~~ \left(\bm{\omega}_{S/T}^T\right)^\top ~~\right]^\top,
\end{equation}
where $\delta\bm{\alpha} \in \mathbb{R}^{6}$ are the osculating Relative Orbital Elements (ROE) parametrizing the 6D state of the target relative to the servicer; $\delta\bm{p} \in \mathbb{R}^{3}$ denotes the MRP vector parametrizing the local error-quaternion; and $\bm{\omega}_{S/T}^T \in \mathbb{R}^{3}$ describes the relative angular velocity of the servicer with respect to the target expressed in the target's principal axes. In this work, a set of nonsingular ROE \cite{Koenig2017STM} is used; however, any representation could be adopted depending on the orbit regime under consideration. The nonsingular ROE are a combination of the equinoctial orbital elements ($\bm{\alpha}$) of both spacecraft as follows,
\begin{equation}
    \bm{\alpha} = \begin{bmatrix} a \\ e_x \\ e_y \\ i_x \\ i_y \\ \lambda  \end{bmatrix} = \begin{bmatrix} a \\ e\cos(\Omega + \omega) \\ e\sin(\Omega + \omega) \\ \tan\big(\frac{i}{2}\big)\cos\Omega \\ \tan\big(\frac{i}{2}\big)\sin\Omega \\ \Omega + \omega + M \end{bmatrix},  \quad\quad
 \delta\bm{\alpha} = \begin{bmatrix} \delta a \\ \delta\lambda \\ \delta e_x \\ \delta e_y \\ \delta i_x \\ \delta i_y \end{bmatrix} = \begin{bmatrix} (a_T - a_S) / a_S \\ \lambda_T - \lambda_S \\ e_{x,T} - e_{x,S} \\ e_{y,T} - e_{y,S} \\ i_{x,T} - i_{x,S} \\ i_{y,T} - i_{y,S} \end{bmatrix},
\end{equation}
where $[a, e, i, \Omega, \omega, M]$ are classical Keplerian orbital elements. The 3D MRP vector is related to a 4D error-quaternion vector $\delta\bm{q} = [\delta q_w ~~ \delta\bm{q}_v^\top]^\top$ via \cite{Tweddle2015MEKF, Sharma2017AASGCC}
\begin{align} \label{eqn:dq to mrp}
	\delta\bm{p} = \frac{4}{1 + \delta q_w} \delta\bm{q}_v.
\end{align}
The factor of 4 ensures that $\|\delta\bm{p}\|$ is approximately equal to the Euler angle for small errors \cite{Crassidis2003UnscentedAttitude}. In USQUE, the propagated MRP state is converted to the error-quaternion via Eq.~\ref{eqn:dq to mrp}, which is then used to update the reference relative quaternion vector, $\bm{q}_{T/S} \in \mathbb{R}^4$, via quaternion multiplication. For more information on the algorithmic implementation of USQUE, readers are referred to \citet{Crassidis2003UnscentedAttitude}.

The time update of the UKF at the $k$-th step propagates the sigma points of the state vector over the propagation interval $\Delta t_k = t_{k} - t_{k-1}$. The advantage of the UKF is that nonlinear dynamics and measurement models can be applied throughout each update, thus retaining higher-order information. For the ROE state, however, a closed-form State Transition Matrix (STM) appropriate for small inter-spacecraft separations in a Keplerian orbit is adopted due to its simplicity, so that
\begin{align} 
	\roe_{k|k-1} = \bm{\Phi}_{\text{NS},k}\left(\bm{\alpha}_{S}(t_{k}), \Delta t_k\right) \roe_{k-1|k-1},
\end{align}
where $\bm{\Phi}_{\text{NS},k}\left(\bm{\alpha}_{S}(t_{k}), \Delta t_k\right)$ is the STM that is a function of the servicer's nonsingular Orbital Elements (OE) at $t_{k}$ and the propagation interval $\Delta t_k$. Note that while a well-defined STM is used for the time update of the ROE state in this work, the UKF framework permits more complex and nonlinear dynamics update procedures for any other relative orbital state representations.

In USQUE, the sigma points for the MRP vector component are converted to quaternion sigma points which are propagated in time. As the MRP vector tracks the error with respect to the reference attitude, it is reset to zero after each time step. Therefore, only the dynamics of the quaternion vector need to be considered, given as
\begin{align} \label{eqn:quaternion dynamics}
	\dot{\bm{q}}_{T/S,k} = \frac{1}{2} \bm{\Omega}\left(\bm{\omega}_{T/S,k}^T\right) \bm{q}_{T/S, k}, ~~\text{where}~~
	\bm{\Omega}(\bm{\omega}) = \begin{bmatrix} 0 & -\bm{\omega}^\top \\ \bm{\omega} & -[\bm{\omega}]_\times \end{bmatrix},
\end{align}
and $[\bm{\omega}]_\times \in \mathbb{R}^{3 \times 3}$ is the skew-symmetric cross product matrix of $\bm{\omega} \in \mathbb{R}^3$. Finally, the expression for relative angular acceleration is derived as \cite{Capuano2018Pose}
\begin{align} \label{eqn:relative ang vel dynamics}
	\dot{\bm{\omega}}_{S/T,k}^T = \bm{R}_{T/S, k} \dot{\bm{\omega}}_{S,k}^S - \bm{I}_T^{-1} \left(\bm{m}_{T,k} - \bm{\omega}_{T,k}^T \times \bm{I}_T \bm{\omega}_{T,k}^T\right) - \bm{\omega}_{T,k}^T \times \bm{\omega}_{S/T, k}^T,
\end{align}
where $\bm{I}_T \in \mathbb{R}^{3 \times 3}$ is  the target's principal moment of inertia matrix, and $\bm{m}_T \in \mathbb{R}^3$ is the control moment of the target about its principal axes. The servicer's absolute angular acceleration, $\dot{\bm{\omega}}_{S,k}^S$, is assumed available from the servicer's Attitude Determination and Control System (ADCS). The target's absolute angular velocity can be computed from current estimates as $\bm{\omega}_{T,k}^T = \bm{R}_{T/S, k}\bm{\omega}_{S,k}^S - \bm{\omega}_{S/T,k}^T$. In this work, the target's inertia matrix is assumed known, and $\bm{m}_T = \bm{0}_{3 \times 1}$ is assumed for a non-operating target spacecraft or debris. In order to accurately update the quaternion and relative angular velocity, and considering the length of the update interval and the rate at which the target could tumble, Equations \ref{eqn:quaternion dynamics} and \ref{eqn:relative ang vel dynamics} are integrated via fourth-order Runge Kutta.

\subsection{UKF Measurement Model} \label{subsec:measurement model}

The measurement vector consists of the ($x$, $y$) pixel coordinates of detected keypoints from $\Hhead$ as well as the regressed translation and rotation vectors from $\Ehead$ of SPNv2. The translation and rotation vectors are denoted as ($\bm{t}_\textrm{E}$, $\bm{q}_\textrm{E}$) respectively. Recall from \mbox{Sec.~\ref{subsec:SPNv2}} that both heatmaps and pose vectors are used as measurements despite their redundancy in order to increase overall robustness. The complete measurement vector, $\bm{y} \in \mathbb{R}^{2K + 7}$, is given as	
\begin{align}
	\bm{y}_k = [~~\bm{y}_{\text{H}, k}^\top ~~ \bm{y}_{\text{E}, k}^\top ~~]^\top = [ ~~ x_{1,k} ~~ y_{1,k} ~~ \dots ~~ x_{K,k} ~~ y_{K,k} ~~ \bm{t}_{\textrm{E}, k}^\top ~~ \bm{q}_{\textrm{E}, k}^\top ~~ ]^\top,
\end{align}
where $\bm{y}_{\text{H}, k} \in \mathbb{R}^{2K}$ and $\bm{y}_{\text{E}, k} \in \mathbb{R}^{7}$ denote measurements from $\Hhead$ and $\Ehead$ respectively .

\subsubsection{Heatmap Measurements}

At the $k$-th step, the modeled measurements for the keypoints can be computed from the current state estimate via projective transformation for a pinhole camera model. Given the camera intrinsic matrix $\bm{M} \in \mathbb{R}^{3 \times 3}$; the known pose of the camera with respect to the servicer spacecraft's principal axes ($\bm{t}_{C/S}^S, \bm{q}_{C/S}$); and the known 3D coordinates of the keypoints in the target's principal frame $\bm{k}_j^T \in \mathbb{R}^{3}, j = 1, \dots, K$, the measurement model for the $j$-th keypoint pixel location is given as
\begin{align}
	\begin{bmatrix} \hat{x}_{j,k} \\ \hat{y}_{j,k} \\ 1 \end{bmatrix} = s\bm{M} [ \bm{R}_{C/T, k} ~|~ \bm{t}_{T/C, k}^C ] \begin{bmatrix} \bm{k}_j^T \\ 1 \end{bmatrix}.
\end{align}
Here, $s$ is an arbitrary scaling factor, and
\begin{align}
	\bm{R}_{C/T, k} &= \bm{R}_{C/S} \bm{R}_{S/T, k}, \\
	\bm{t}_{T/C, k}^C &= \bm{R}_{C/S} \bm{t}_{T/S, k}^S + \bm{t}_{S/C}^C.
\end{align}

\begin{figure}[!t]
	\centering
	\includegraphics[width=0.75\textwidth]{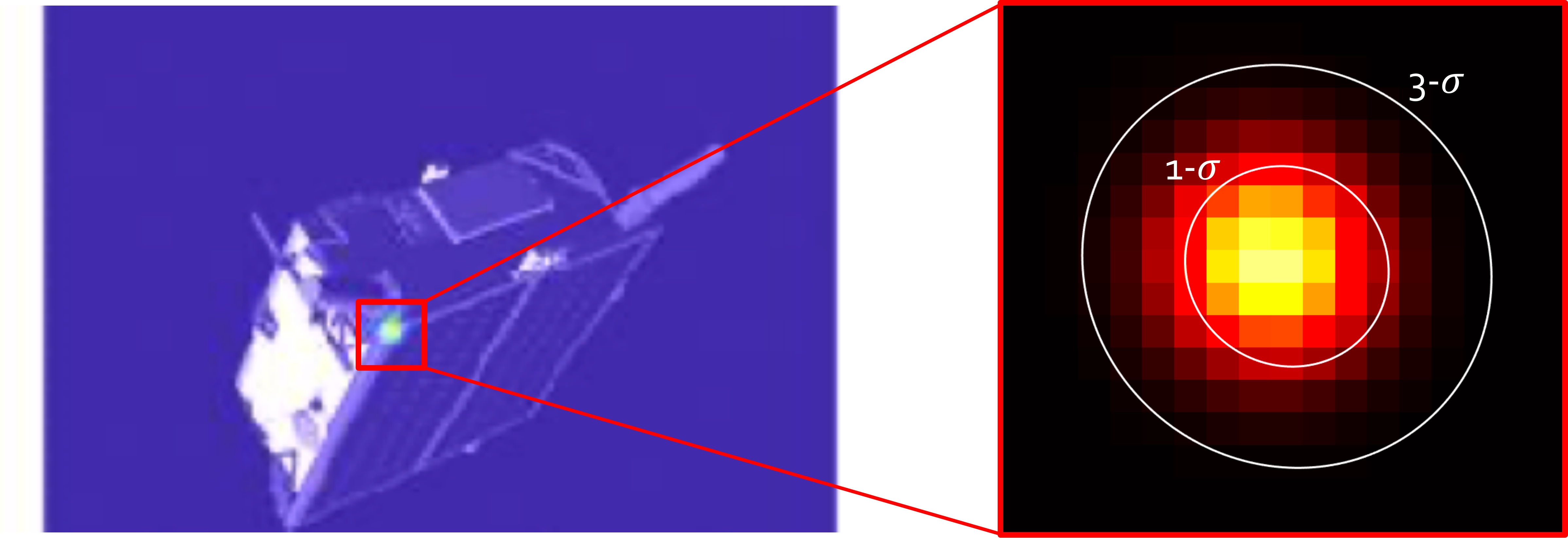}
	\caption{Detected heatmap about a keypoint and its spread as a Gaussian distribution.}
	\label{fig:heatmap}
\end{figure}

The spread of the heatmap about its peak can be interpreted as a confidence associated with the prediction of the keypoint location (see Fig.~\ref{fig:heatmap}). Thus, the covariance matrix associated with the $(x, y)$-coordinates of an $i$-th keypoint, $\bm{C}_\textrm{H}^{(i)} \in \mathbb{R}^{2 \times 2}$, can be computed as \cite{PasqualettoCassinis2021Coupled}
\begin{align}
	\bm{C}_\textrm{H}^{(i)} = \begin{bmatrix} cov(x, x) & cov(x, y) \\ cov(y, x) & cov(y, y) \end{bmatrix}, ~~\text{where}~~ cov(x, y) = \sum_{j=1}^{P} w_j (x_j - p_x) (y_j - p_y).
\end{align}
Here, $(p_x, p_y)$ denotes the coordinates of the peak, $w_j$ is the normalized intensity of the $j$-th pixel, and $P$ is the number of pixels in the image. Then, the covariance matrices for each keypoint are used to construct the corresponding portion of the measurement noise matrix $\bm{R} \in \mathbb{R}^{(2K+6) \times (2K+6)}$ at each iteration. This is performed by populating the $2 \times 2$ entries along the diagonal of the upper-left $2K \times 2K$ portion of $\bm{R}$ with the corresponding covariance matrices $\bm{C}_\text{H}$. For more details, readers are referred to \citet{PasqualettoCassinis2021Coupled}.

\subsubsection{Vector Measurements} \label{subsubsec:vector measurements}

For the vector measurements regressed from $\Ehead$ of SPNv2, the modeled translation vector ($\hat{\bm{t}}_\text{E}$) can be converted from the nonsingular ROE state $\delta\bm{\alpha}_{k|k-1}$, and the modeled quaternion vector ($\hat{\bm{q}}_\text{E}$) is simply mapped from the current estimate of the quaternion state, $\bm{q}_{T/S, k|k-1}$. However, note that quaternion vectors are subject to a unit norm constraint which prohibits computation of measurement residuals by simple subtraction and may cause corresponding $4 \times 4$ measurement covariance matrices to become singular. Therefore, this work borrows techniques from USQUE \citep{Crassidis2003UnscentedAttitude} to properly handle the quaternion vector measurement and its associated covariance matrix in the UKF. The quaternion measurement innovation (i.e., the difference between the observed and expected measurements) is expressed as an MRP vector by first computing the quaternion difference $\delta\bm{q}_\text{E} = \bm{q}_\text{E} \otimes \hat{\bm{q}}_\text{E}^{-1}$, where $\otimes$ denotes quaternion multiplication. $\delta\bm{q}_\text{E}$ is then converted to MRP via Eq.~\ref{eqn:dq to mrp}. 

Note that MRP has a singularity at $360^\circ$, and the difference between the observed and modeled quaternion can be arbitrarily large. To avoid nearing the singularity, the conversion of the quaternion measurement innovation to MRP computes both regular and shadow MRP vectors. The shadow MRP, defined as $\delta\bm{p}^S = 4\delta\bm{q}_{v}/\left(\delta q_{w} - 1\right)$, denotes the same attitude as $\delta\bm{p}$ in Eq.~\ref{eqn:dq to mrp} due to the bijective nature of mapping from quaternions to MRP; however, it has a singularity at $0^\circ$ instead \citep{Markley2014SADCTextbook}. Therefore, by choosing the MRP vector with a smaller norm, one can avoid the singularity of the MRP vector for the quaternion measurement innovation. This MRP innovation is then used to compute the measurement covariance matrix and Kalman gain in the UKF.

Since $\Ehead$ of SPNv2 only outputs the regressed vector measurements, the measurement noise covariance for the $\Ehead$ measurements cannot be estimated online on an unknown image domain as is done for the keypoints. Operationally, since the spaceborne images are unavailable during the on-ground validation phase, the lower-right $6 \times 6$ portion of $\bm{R}$ (denoted as $\bm{C}_\text{E}$) could instead be estimated from the CNN's performance on images obtained from a HIL robotic facility such as TRON. In this work, TRON acts as the actual operational environment and cannot be used to aid the filter, so $\bm{C}_\text{E}$ is derived from SPNv2 performance on the $\synthetic$ validation set of SPEED+. More details are provided in Sec.~\ref{section:experiments}.

\subsection{Outlier Rejection} \label{subsec:outlier rejection}

To mitigate unexpected failures of SPNv2 on test domain images that were unseen during its offline training phase, any outlier measurements are detected and discarded based on the squared Mahalanobis distance of the UKF innovation \cite{Tweddle2015MEKF}, defined as
\begin{align} \label{eqn:innovation}
	d_k^2 = \bm{\Delta}_k^{y \top} \bm{S}_k^{-1} \bm{\Delta}_k^y,
\end{align}
where $\bm{S}_k$ is the measurement covariance matrix of UKF, and $\bm{\Delta}_k^y$ is the innovation or pre-fit residual defined as 
\begin{align} \label{eqn:prefit_residual}
	\bm{\Delta}_k^y = \bm{y}_k - \bm{h}(\bm{x}_{k|k-1}).
\end{align}
Here, $\bm{h}(\cdot)$ is the nonlinear measurement model, and $\bm{x}_{i|i-1}$ is the a priori state estimate at the $k$-th step. For the quaternion vector measurements, the MRP vector corresponding to the quaternion difference is used as explained in Sec.~\ref{subsubsec:vector measurements}. The squared Mahalanobis distance is computed for each keypoint, translation vector, and MRP vector. Since $d^2$ follows the Chi-Square distribution with 2 Degrees-of-Freedom (DoF) for keypoints and 3 DoF for translation and MRP vectors, the corresponding measurement is rejected if it is beyond the threshold determined by the inverse Chi-Square distribution at some specified probability $p$. In this work, $p = 0.99$ is set. If all measurements are rejected, only the time update is performed.

\subsection{Adaptive State Noise Compensation} \label{subsec:ASNC}

Adaptive State Noise Compensation (ASNC) \cite{Stacey2021ASNC} adaptively tunes the process noise covariance matrix $\bm{Q} \in \mathbb{R}^{n \times n}$ at each time step. ASNC ensures that the tuned process noise matrix is positive semi-definite while respecting the continuous time-varying dynamics model of the system. Ordinary State Noise Compensation (SNC) models the process noise covariance at time step $k$ as
\begin{align} \label{eqn:Q_eq}
	\bm{Q}_k = \int_{t_{k-1}}^{t_k} \bm{\Phi}(t_k, \tau)\bm{\Gamma}(\tau)\bm{\widetilde{Q}}_k\bm{\Gamma}(\tau)^\top \bm{\Phi}(t_k, \tau)^\top d\tau,
\end{align}
where $\bm{\Phi}(t_k, t)$ is the STM which propagates the state vector from time $t$ to $t_k$, $\bm{\Gamma}(t)$ is the process noise mapping matrix, and $\bm{\widetilde{Q}}_k$ is the process noise power spectral density matrix. The $\bm{\widetilde{Q}}_k$ matrix is assumed constant over the measurement interval; moreover, the process noise is assumed independent across dimensions such that $\bm{\widetilde{Q}}_k$ is diagonal. Then, Eq.~\ref{eqn:Q_eq} becomes linear in $\bm{\widetilde{Q}}_k$, and the unique elements of the symmetric matrix $\bm{Q}_k$ and the diagonal elements of $\bm{\widetilde{Q}}_k$ can be related as
\begin{align} \label{eqn:ASNC linear mapping}
	\bm{Q}_k^\text{vech} = \bm{X}_k \bm{\widetilde{Q}}_k^\text{diag},
\end{align}
Here, $\bm{A}^\text{vech} = \text{vech}(\bm{A})$ denotes the half-vectorization operation which returns a vector of the lower-triangular elements of the symmetric matrix $\bm{A}$, and $\bm{A}^\text{diag} = \text{diag}(\bm{A})$ returns a vector of the diagonal elements of $\bm{A}$. The linear mapping matrix $\bm{X}_k$ is based on $\bm{\Phi}$ and $\bm{\Gamma}$ which vary depending on the state representation and the underlying dynamics model. 

In SNC, the diagonal matrix $\bm{\widetilde{Q}}_k$ is manually tuned offline. ASNC instead solves for the optimal $\bm{\widetilde{Q}}_k$ by matching Eq.~\ref{eqn:Q_eq} with the corresponding estimate $\bm{\hat{Q}}_k$ obtained through covariance matching over a sliding window \cite{Fraser2021AdaptiveKalman}, i.e., 
\begin{align} \label{eqn:covariance matching}
	\bm{\hat{Q}}_k = \frac{1}{N_W} \sum_{i = k-N_W+1}^k \bm{P}_{i|i} - \bm{\Phi}_i\bm{P}_{i-1|i-1}\bm{\Phi}_i^\top +  \bm{\Delta}_i^x \bm{\Delta}_i^x{}^\top
\end{align}
where $\bm{P}_{i|i}$ is the a posteriori state covariance matrix at the $i$-th step. Here, $\bm{\Delta}_i^x$ is the state correction term defined as
\begin{align}
	\bm{\Delta}_i^x = \bm{K}_k \bm{\Delta}_i^y,
\end{align}
where $\bm{K}_k$ is the Kalman gain, and $\bm{\Delta}_i^y$ is the pre-fit residual in Eq.~\ref{eqn:prefit_residual}.
 
The optimal $\bm{\widetilde{Q}}_k$ for Eq.~\ref{eqn:Q_eq} is then the solution to the constrained weighted least-squares minimization problem
\begin{equation} \label{eqn:ASNC LS minimizaiton}
	\begin{split}
		\min_{\bm{\widetilde{Q}}^\text{diag}} &~~ (\bm{X}_k \bm{\widetilde{Q}}^\text{diag} - \bm{\hat{Q}}_{k+1}^\text{vech})^\top \bm{W}_{k+1}^{-1}(\bm{X}_k \bm{\widetilde{Q}}^\text{diag} - \bm{\hat{Q}}_{k+1}^\text{vech}) \\
		\text{subject to} &~~ \bm{\widetilde{Q}}_\ell^\text{diag} \leq \bm{\widetilde{Q}}^\text{diag} \leq \bm{\widetilde{Q}}_u^\text{diag},
	\end{split}
\end{equation}
where $\bm{W}_{k+1}$ is the theoretical covariance of $\hat{\bm{Q}}_{k+1}^\text{vech}$, and $\bm{\widetilde{Q}}_\ell^\text{diag}$ and $\bm{\widetilde{Q}}_u^\text{diag}$ are respectively the element-wise lower- and upper-bounds on $\bm{\widetilde{Q}}_\text{diag}$ based on coarse a priori knowledge of the dynamical environment. For details regarding the solving of Eq.~\ref{eqn:ASNC LS minimizaiton}, readers are referred to \citet{Stacey2021ASNC}.

\citet{Stacey2022CovModel} derived analytical process noise models for various absolute and relative orbital state representations by assuming two-body motion and that the noise manifests as unmodeled acceleration in the Radial-Tangential-Normal (RTN) frame. The next section discusses a process noise model for the nonsingular ROE representation at small separations and additionally derives a new analytical process noise model for relative attitude motion assuming a slow tumbling rate of the target.

\section{Analytical Process Noise Models} \label{sec:Process Noise Modeling}

In order to perform ASNC to adaptively tune the process noise covariance matrix, the process noise model must be derived for the state vector described in Eq.~\ref{eqn:state_vector}. In this work, the noise components of the orbital and attitude states are decoupled to facilitate the derivation and computation, i.e., 
\begin{align}
	\bm{Q}_k = \begin{bmatrix} \bm{Q}_{\roe, k} & \bm{0}_{6 \times 6} \\ \bm{0}_{6 \times 6} & \bm{Q}_{\bm{q}, k} \end{bmatrix}.
\end{align}
The process noise power spectral density matrix for unmodeled relative Cartesian accelerations is modeled in the servicer's RTN frame, whereas that of the unmodeled differential angular accelerations is modeled about the target's principal axes. For simplicity, the process noise power spectral density matrices for both relative orbit and attitude states are assumed diagonal, i.e.,
\begin{align}
	\bm{\widetilde{Q}}_{\roe}= \begin{bmatrix} \widetilde{Q}_{\roe}^r & 0 & 0 \\ 0 & \widetilde{Q}_{\roe}^t & 0 \\ 0 & 0 & \widetilde{Q}_{\roe}^n \end{bmatrix}, ~~ \bm{\widetilde{Q}}_{\bm{q}} = \begin{bmatrix} \widetilde{Q}_{\bm{q}}^x & 0 & 0 \\ 0 & \widetilde{Q}_{\bm{q}}^y & 0 \\ 0 & 0 & \widetilde{Q}_{\bm{q}}^z \end{bmatrix}.
\end{align}
Then, taking the ROE state as an example, the process noise covariance matrix in Eq.~\ref{eqn:Q_eq} becomes
\begin{align} \label{eqn:Q_roe as sum of diagonal}
	\bm{Q}_{\roe, k} = \bm{X}_{k}^r\widetilde{Q}_{\roe,k}^r + \bm{X}_{k}^t\widetilde{Q}_{\roe,k}^t + \bm{X}_{k}^n\widetilde{Q}_{\roe,k}^n,
\end{align}
where 
\begin{align} \label{eqn:X matrix def - general}
	\bm{X}_{k}^i = \int_{t_{k-1}}^{t_k} \bm{\bar{\Gamma}}_{k}^i(t_k ,\tau) \bm{\bar{\Gamma}}_{k}^i(t_k ,\tau)^\top d\tau, ~~~ i \in \{r, t, n\}.
\end{align}
Here, $\bm{\bar{\Gamma}}_k(t_k, t) = [\bm{\bar{\Gamma}}_{k}^r$~~~$\bm{\bar{\Gamma}}_{k}^t$~~~$\bm{\bar{\Gamma}}_{k}^n] = \bm{\Phi}(t_k, t) \bm{\Gamma}_k(t)$. Equations \ref{eqn:Q_roe as sum of diagonal} and \ref{eqn:X matrix def - general} can now be used to construct the linear mapping of Eq.~\ref{eqn:ASNC linear mapping} as
\begin{align} \label{eqn:ASNC Linear Mapping Expanded}
	\bm{Q}_{\delta\bm{\alpha}, k}^\text{vech} = \bm{X}_k \bm{\widetilde{Q}}_{\roe,k}^\text{diag} = \begin{bmatrix} \vert & \vert & \vert \\ \text{vech}(\bm{X}_{k}^r) & \text{vech}(\bm{X}_{k}^t) & \text{vech}(\bm{X}_{k}^n) \\ \vert & \vert & \vert \end{bmatrix} \begin{bmatrix} \widetilde{Q}_{\roe,k}^r \\ \widetilde{Q}_{\roe,k}^t \\ \widetilde{Q}_{\roe,k}^n \end{bmatrix}.
\end{align}
A similar expression can be constructed for the attitude dynamics. Once $\bm{X}_k$ matrices can be constructed from Eq.~\ref{eqn:ASNC Linear Mapping Expanded} for both the ROE and attitude states, the weighted least-squares minimization problem of Eq.~\ref{eqn:ASNC LS minimizaiton} can be solved individually for both states using an off-the-shelf least-squares or quadratic programming solver. In this work, MATLAB's \texttt{lsqlin} command is used to solve Eq.~\ref{eqn:ASNC LS minimizaiton} with a non-negativity constraint, i.e., $\bm{\widetilde{Q}}_\ell^\text{diag} = \bm{0}_{3 \times 1}$, to ensure a positive semi-definite solution.

The following sections describe the analytical process noise covariance models for both states. The derivation of the model for the attitude motion is a new and essential contribution of this work.

\subsection{ROE State Process Noise}

For a nonsingular ROE representation based on the equinoctial elements, \citet{Stacey2022CovModel} derived the process noise covariance model under the assumption of two-body motion. Specifically, for a small separation between two spacecraft, the authors first derive the process noise $\bm{Q}_{\roe^\prime}$ for an alternative ROE representation defined as $\roe^\prime = \bm{\alpha}_T - \bm{\alpha}_S$. Then, the process noise for nonsingular ROE can be recovered via
\begin{align} \label{eqn:Q from OE diff to NS ROE}
	\bm{Q}_{\delta\bm{\alpha}, k} = \bm{J}_{\delta\bm{\alpha}}(t_k) \bm{Q}_{\delta\bm{\alpha}^\prime, k} \bm{J}_{\delta\bm{\alpha}}(t_k)^\top
\end{align}
where 
\begin{align} \label{eqn:J from OE diff to NS ROE}
	\bm{J}_{\delta\bm{\alpha}}(t_k) = \frac{\partial \delta\bm{\alpha}}{\partial \delta\bm{\alpha}^\prime} \bigg|_{\delta\bm{\alpha}^\prime = \bm{0}} = 
	\begin{bmatrix} \frac{1}{a_{S,k}} & \bm{0}_{1 \times 4} & 0 \\ 0 & \bm{0}_{1 \times 4} & 1 \\ \bm{0}_{4 \times 1} & \textbf{I}_{4 \times 4} & \bm{0}_{4 \times 1} \end{bmatrix}
\end{align}
and $a_{S,k}$ is the semi-major axis of the servicer at $t_k$. Noting that $\bm{Q}_{\roe^\prime} = \sum_{i \in \{r, t, n\}} \bm{X}_{k}^{i} {}^\prime \widetilde{Q}_{\roe^\prime}$ as in Eq.~\ref{eqn:Q_roe as sum of diagonal}, the linear mapping matrices of Eq.~\ref{eqn:ASNC Linear Mapping Expanded} are now given as
\begin{align}
	\bm{X}_{k}^i =  \bm{J}_{\delta\bm{\alpha}}(t_k) \bm{X}_{k}^{i} {}^\prime  \bm{J}_{\delta\bm{\alpha}}(t_k)^\top
\end{align}
where $\bm{X}_{k}^{i} {}^\prime$ of Eq.~\ref{eqn:X matrix def - general} for $\roe^\prime$ is derived by \citet{Stacey2022CovModel} and partially reproduced in Appendix \ref{appendix:Process Noise Models ROE}.

\subsection{Attitude State Process Noise}

In order to derive the process noise model for the attitude states, the STM ($\bm{\Phi}_{\bm{q}, k}$) and the process noise mapping matrix ($\bm{\Gamma}_k$) must first be constructed. The kinematics of the MRP vector for attitude error are \cite{Younes2019AttErrDynamics, Markley2003AttErrRep}
\begin{align}
	\delta\dot{\bm{p}} &= \bigg[-\frac{1}{2} \big([\delta\bm{\omega}]_\times + 2[\hat{\bm{\omega}}]_\times \big) + \frac{1}{8} \delta\bm{\omega}^\top \delta\bm{p}\bigg] \delta\bm{p} + \bigg[ 1 - \frac{1}{16}\delta\bm{p}^\top \delta\bm{p}\bigg] \delta\bm{\omega},
\end{align}
where a shorthand notation of $\bm{\omega} \equiv \bm{\omega}_{T/S}^T$ is used, and $\delta\bm{\omega} \equiv \bm{\omega} - \hat{\bm{\omega}}$, where the hat denotes an estimated quantity. In the MEKF and USQUE frameworks, the MRP vector corresponding to the error-quaternion state is reset to zero prior to each propagation step. Therefore, assuming short propagation intervals and a small relative angular velocity, the MRP kinematics equation simplifies to \cite{Crassidis2003UnscentedAttitude}
\begin{align} \label{eqn:mrp dynamics - approx}
	\delta\dot{\bm{p}} \approx  -\frac{1}{2} \big( [\delta\bm{\omega}]_\times + 2[\hat{\bm{\omega}}]_\times \big) \delta\bm{p} + \delta\bm{\omega},
\end{align} 
where the higher-order terms $\delta\bm{\omega}^\top \delta\bm{p}$ and $\delta\bm{p}^\top \delta\bm{p}$ are assumed negligible. Then, to derive the STM, Eq.~\ref{eqn:mrp dynamics - approx} is linearized about its attitude state elements, ($\delta\bm{p}, \bm{\omega}_{S/T}^T$), as noted in Eq.~\ref{eqn:state_vector}. Specifically, the linearization is performed at $\delta\bm{p} = \bm{0}$, $\bm{\omega} = \hat{\bm{\omega}}$, which results in
\begin{align} \label{eqn:mrp dynamics - linearized}
    \delta\dot{\bm{p}} = -[\hat{\bm{\omega}}]_\times \delta\bm{p} + \bm{\omega} = [\hat{\bm{\omega}}_{S/T}^T]_\times \delta\bm{p} - \bm{\omega}_{S/T}^T.
\end{align}

\noindent Likewise, the relative angular velocity dynamics in Eq.~\ref{eqn:relative ang vel dynamics} approximate to 
\begin{align} \label{eqn:rel ang vel dynamics - approx}
	\dot{\bm{\omega}}_{S/T}^T \approx -[\bm{R}_{T/S}\bm{\omega}_S^S]_\times \bm{\omega}_{S/T}^T - \bm{I}_T^{-1} \bm{\eta}_T + \bm{R}_{T/S}\dot{\bm{\omega}}_S^S,
\end{align}
where the gyroscopic term $\bm{\omega}_T^T \times \bm{I}_T \bm{\omega}_T^T$ is assumed negligible, which is a reasonable assumption for a small spin rate and exact if the target spins about one axis. In Eq.~\ref{eqn:rel ang vel dynamics - approx}, $\bm{\eta}_T \in \mathbb{R}^3$ accounts for any unmodeled torques in the system expressed in the target's principal axes frame. The continuous-time dynamics can now be constructed from Eqs.~\ref{eqn:mrp dynamics - linearized}, \ref{eqn:rel ang vel dynamics - approx} as
\begin{gather}
\begin{aligned}
	&\begin{bmatrix} \delta\dot{\bm{p}} \\ \dot{\bm{\omega}}_{S/T}^T \end{bmatrix} = 
    \bm{A} \begin{bmatrix} \delta\bm{p} \\ \bm{\omega}_{S/T}^T \end{bmatrix}  + \bm{\Gamma} \bm{\eta}_T + \bm{B}, \\ 
    \text{where} \quad &
    \bm{A} = \begin{bmatrix} [\hat{\bm{\omega}}_{S/T}^T]_\times & -\bm{I}_{3 \times 3} \\ \bm{0}_{3 \times 3} & -[\bm{R}_{T/S}\bm{\omega}_S^S]_\times \end{bmatrix}, \quad
    \bm{\Gamma} = \begin{bmatrix} \bm{0}_{3 \times 3} \\ -\bm{I}_T^{-1} \end{bmatrix}, \quad
    \bm{B} = \begin{bmatrix} \bm{0}_{3 \times 1} \\ \bm{R}_{T/S}\dot{\bm{\omega}}_S\end{bmatrix}.
\end{aligned}
\end{gather}
Here, $\bm{A}$ is the plant matrix, and $\bm{\Gamma}$ is the process noise mapping matrix. The $\bm{B}$ matrix contains the absolute angular acceleration of the servicer due to its natural dynamics and potential attitude control maneuvers. This information is assumed available from the servicer's ADCS. Then, the following STM can be obtained via zero-hold integration \cite{Tweddle2015MEKF},
\begin{align}
	\bm{\Phi}_{\bm{q}, k}(t,0) = \begin{bmatrix} e^{[\hat{\bm{\omega}}_{S/T, k}^T]_\times t} & -\int_{0}^{t} e^{[\hat{\bm{\omega}}_{S/T, k}^T]_\times \tau} d\tau \\ \bm{0}_{3 \times 3} & e^{-[\bm{R}_{T/S, k}{\bm{\omega}_{S,k}^S]_\times t}} \end{bmatrix},
\end{align}
which leads to 
\begin{align} \label{eqn:Gamma_bar for attitude, raw with matrix exp}
	\bm{\bar{\Gamma}}_k(t, 0) = \bm{\Phi}_{\bm{q}, k}(t, 0) \bm{\Gamma}_k = \begin{bmatrix} \int_{0}^{t} e^{[\hat{\bm{\omega}}_{S/T, k}^T]_\times \tau} d\tau \\ -e^{-[\bm{R}_{T/S, k}\bm{\omega}_{S,k}^S]_\times t}  \end{bmatrix} \bm{I}_T^{-1} = \begin{bmatrix} \bm{\Lambda}_1(t) \\ \bm{\Lambda}_2(t) \end{bmatrix} \bm{I}_T^{-1}.
\end{align}
In order for $\bar{\bm{\Gamma}}_k$ to be used in Eq.~\ref{eqn:X matrix def - general} to compute the linear mapping matrix $\bm{X}_{k}^i$ for $i \in \{x, y, z\}$, the integral of the matrix exponential must be evaluated. From Rodrigues' formula, the exponential of a real, skew-symmetric matrix $\bm{\Theta} = [\bm{\theta}]_\times \in \mathbb{R}^{3 \times 3}$ is given as
\begin{align}
	e^{\bm{\Theta}} = \textbf{I}_{3 \times 3} + \sin \theta \hat{\bm{\Theta}} + (1 - \cos \theta) \hat{\bm{\Theta}}^2,
\end{align}
where $\theta = \| \bm{\theta} \|$ and $\bm{\hat{\Theta}} = \bm{\Theta} / \theta$. Applying this to the integrand of the integral term in Eq.~\ref{eqn:Gamma_bar for attitude, raw with matrix exp} yields
\begin{align}
	e^{[\bm{\omega}_1]_\times \tau} = \textbf{I}_{3 \times 3} + \sin \omega_1\tau [\hat{\bm{\omega}}_1]_\times + ( 1 - \cos \omega_1\tau )[\hat{\bm{\omega}}_1]_\times^2,
\end{align}
where $\bm{\omega}_1$ is a shorthand notation for $\bm{\omega}_{S/T, k}^T$, $\omega_1 = \| \bm{\omega}_1 \|$, and $[\hat{\bm{\omega}}_1]_\times = [\bm{\omega}_1]_\times / \omega_1$. Integrating over $[0, t]$, one obtains
\begin{align}
	\bm{\Lambda}_1(t) = \int_0^t e^{[\bm{\omega}_1]_\times \tau} d\tau = \textbf{I}_{3 \times 3}t + \frac{1}{\omega_1}( 1 - \cos\omega_1t) [\hat{\bm{\omega}}_1]_\times + (t - \frac{1}{\omega_1} \sin\omega_1 t ) [\hat{\bm{\omega}}_1]_\times^2.
\end{align}
Likewise, 
\begin{align}
	\bm{\Lambda}_2(t) = -e^{-[\bm{\omega}_2]_\times t} = -\bigg( \textbf{I}_{3 \times 3} - \sin \omega_2 t [\bm{\hat{\omega}}_2]_\times + ( 1 - \cos \omega_2 t) [\bm{\hat{\omega}}_2]_\times^2 \bigg),
\end{align}
where $\bm{\omega}_2$ is a shorthand notation for $\bm{R}_{T/S, k}\bm{\omega}_{S,k}^S$. The $\bm{X}_{k}^i$ matrix in Eq.~\ref{eqn:X matrix def - general} can now be computed as
\begin{align} \label{eqn:X matrix for attitude in A, B, C}
	\bm{X}_{k}^i = \bm{I}_{T,i}^{-2} \begin{bmatrix} \bar{\bm{A}}_i  & \bar{\bm{B}}_i \\ \bar{\bm{B}}_i^\top & \bar{\bm{C}}_i \end{bmatrix} 
\end{align}
where $\bm{I}_{T,i}$ is the $i$-th diagonal element of $\bm{I}_T$. The analytical expressions for sub-matrices $\bar{\bm{A}}_i, \bar{\bm{B}}_i, \bar{\bm{C}}_i \in \mathbb{R}^{3 \times 3}$ are provided in Appendix \ref{appendix:Process Noise Models Attitude}.

In summary, the above formulations allow for accurate modeling of the process noise covariance matrix associated with relative attitude motion based on the underlying continuous dynamics of the sytem. The assumptions are representative and benign for the purpose of deriving analytical expressions which enable the least-squares minimization of ASNC in Eq.~\ref{eqn:ASNC LS minimizaiton}.
\section{Satellite Hardware-In-the-loop Rendezvous Trajectories (SHIRT) Dataset} \label{sec:SHIRT}

Since the proposed UKF pipeline incorporates a CNN model trained on synthetic images, the filter's performance must also be evaluated on images captured from a real-life source. The goal is to demonstrate that the filter displays low steady-state errors despite the domain gap experienced by SPNv2. Inspired by the HIL images of the authors' SPEED+ dataset \citep{Park2021speedplus}, and in order to harness the ability to generate real images for any desired trajectory, this work introduces the Satellite Hardware-in-the-loop Rendezvous Trajectories (SHIRT) dataset, which consists of HIL images of a known target captured in two simulated rendezvous trajectories in LEO.

\subsection{Reference Trajectory Simulation} \label{section:shirt:trajsim}

\begin{table*}[!t]
	\caption{Initial mean absolute orbital elements of the servicer and relative orbit elements of the target with respect to the servicer.}
	\label{tab:SHIRT initial orbit}
	\centering
	\tabcolsep=0.1cm
	\begin{tabular}{lcccccccccccc}
		\toprule
		& \multicolumn{6}{c}{\bfseries Servicer Mean OE} & \multicolumn{6}{c}{\bfseries Target Mean ROE} \\
		\cmidrule(lr){2-7} \cmidrule(lr){8-13}
		& $a$ [km] & $e$ [-] & $i$ [${}^\circ$] & $\Omega$ [${}^\circ$] & $\omega$ [${}^\circ$] & $M$ [${}^\circ$] & $a\delta a$ [m] & $a\delta\lambda$ [m] & $a\delta e_x$ [m] & $a\delta e_y$ [m] &  $a\delta i_x$ [m] & $a\delta i_y$ [m] \\
		\midrule
		ROE1 & \multirow{2}{*}{7078.135} & \multirow{2}{*}{0.001} & \multirow{2}{*}{98.2} & \multirow{2}{*}{189.9} & \multirow{2}{*}{0} & \multirow{2}{*}{0} & 0 & -8 & 0 & 0 & 0 & 0 \\
		ROE2 & &&&&&& -0.250 & -8.1732 & 0.0257 & -0.1476 & -0.030 & 0.1724 \\
		\bottomrule
	\end{tabular}
\end{table*}

Drawing inspiration from \citet{Sharma2017AASGCC} and \citet{Damico2010PHDThesis}, SHIRT includes simulations of two reference trajectories that emulate typical rendezvous scenarios in LEO. The initial mean absolute OE of the servicer and ROE of the target with respect to the servicer are presented in Table \ref{tab:SHIRT initial orbit}. The ROE1 scenario maintains an along-track separation typical of a standard v-bar hold point, whereas ROE2 introduces a small, nonzero relative semi-major axis ($\delta a$) so that the servicer slowly approaches the target. The servicer's initial mean OE, derived from the PRISMA mission \cite{PRISMA_chapter, Damico2014IJSSE}, indicate a dawn-dusk sun-synchronous orbit with 18 h nominal Local Time at the Ascending Node (LTAN). 

The servicer's initial attitude, which coincides with the camera's attitude, is defined with respect to the RTN frame. Specifically, the camera boresight (i.e., $z$-axis) is initially directed along the negative along-track direction ($-\hat{T}$) and its $x$-axis along the cross-track direction ($\hat{N}$). The servicer's attitude is controlled such that the camera boresight is always directed along $-\hat{T}$. Moreover, the servicer's angular velocity about its body axes is set to [$n$ ~~ 0 ~~ 0]$^\top$ (rad/s), where $n$ is the satellite mean motion, and torque is applied at each time step to negate any accumulated environmental perturbation moments. The target's initial relative attitude with respect to the servicer is given in terms of a quaternion as $\bm{q}_o = [1/\sqrt{2} ~~ 1/\sqrt{2} ~~0 ~~ 0]^\top$. The target's initial angular velocity about its principal axes is set to $\bm{\omega}_0$ = [1 ~~ 0 ~~ 0]$^\top$ (${}^\circ$/s) for ROE1 and $\bm{\omega}_0$ = [0 ~~ 0.4 ~~ -0.6]$^\top$ (${}^\circ$/s) for ROE2, which is reasonable for a tumbling, non-cooperative RSO. Based on these conditions, the target's $x$-axis is initially aligned with the cross-track direction. Since the target rotates about its $x$-axis only in ROE1, and since the servicer maintains a nearly constant separation from the target, the servicer's camera observes the target from a limited range of viewpoints. ROE1 is therefore a much more difficult scenario than ROE2 because it observes greater geometric dilution of precision.

\begin{table*}[!t]
	\caption{SHIRT simulation parameters.}
	\label{tab:SHIRT simulation parameters}
	\centering
	\begin{tabular}{ll}
		\toprule
		\bfseries \underline{Simulation Parameters} & \\
		Initial epoch & 2011/07/18 01:00:00 UTC \\
		Integrator & Runge-Kutta (Dormand-Prince) \cite{Dormand1980DOPRI}\\
		Step size  & 1 s\\
		Simulation time & 2 orbits (3.3 hrs) \\
		\midrule
		\bfseries \underline{Force Models} & \\
		Geopotential field (degree $\times$ order) & GGM05S (120 $\times$ 120) \cite{Ries2016GGM05} \\
		Atmospheric density & NRLMSISE-00 \cite{Picone2002NRLMSISE00} \\
		Solar radiation pressure & Cannon-ball, conical Earth shadow \\
		Third-body gravity & Analytical Sun \& Moon \cite{Montenbruck2013SatelliteOrbits} \\
		Relativistic effect & 1st order \cite{Montenbruck2013SatelliteOrbits} \\
		\midrule
		\bfseries \underline{Torque Models} & \\
		Gravity gradient & Analytical \cite{Wertz1978ADCS} \\
		Atmospheric density & NRLMSISE-00 \cite{Picone2002NRLMSISE00} \\
		Solar radiation pressure & Conical Earth shadow \\
		Geomagnetic field (order) & IGRF-13 (10) \cite{Alken2020IGRF-13} \\
		\bottomrule
	\end{tabular}
\end{table*}

\begin{table*}[!t]
	\caption{Spacecraft parameters of Mango (servicer) and Tango (target) of the PRISMA mission \cite{PRISMA_chapter} for force and torque models evaluation.}
	\label{tab:mango_tango_parameters}
	\centering
	\begin{tabular}{lll}
		\toprule
		\bfseries Spacecraft Parameters & \bfseries Servicer (Mango) &  \bfseries Target (Tango) \\
		\midrule
		\bfseries \underline{Force Model Evaluation} & & \\
		Spacecraft mass [kg] & 154.4 & 42.5 \\
		Cross-sectional area (drag) [m$^2$] & 1.3 & 0.38 \\
		Cross-sectional area (SRP) [m$^2$] & 2.5 & 0.55 \\
		Aerodynamic drag coefficient & 2.5 & 2.25 \\
		SRP coefficient & 1.32 & 1.2 \\
		\midrule
		\bfseries \underline{Torque Model Evaluation} & & \\
		Number of faces & 10 & 6 \\
		Principal moment of inertia [kg$\cdot$m$^2$] & $\text{diag}(16.70, 19.44, 18.28)$ & $\text{diag}(2.69, 3.46, 3.11)$ \\
		\makecell[l]{Direction Cosine Matrix (DCM) \\ from body to principal frame} & $\begin{bmatrix} 1 &  0  & 0 \\ 0 & 1 & 0 \\ 0 & 0 & 1 \end{bmatrix}$ & $\begin{bmatrix} 1 &  0  & 0 \\ 0  & -0.929  &  0.369 \\ 0 &  -0.369 &  -0.929 \end{bmatrix}$ \\
		Magnetic dipole moment [A$\cdot$m$^2$] & $[0, 0, 0]^\top$ & $[0, 0, 5.66\times 10^{-7}]^\top$ \\
		\bottomrule
	\end{tabular}
\end{table*}

\begin{figure}[!t]
	\centering
	\includegraphics[width=0.9\textwidth, trim=4 4 4 4,clip]{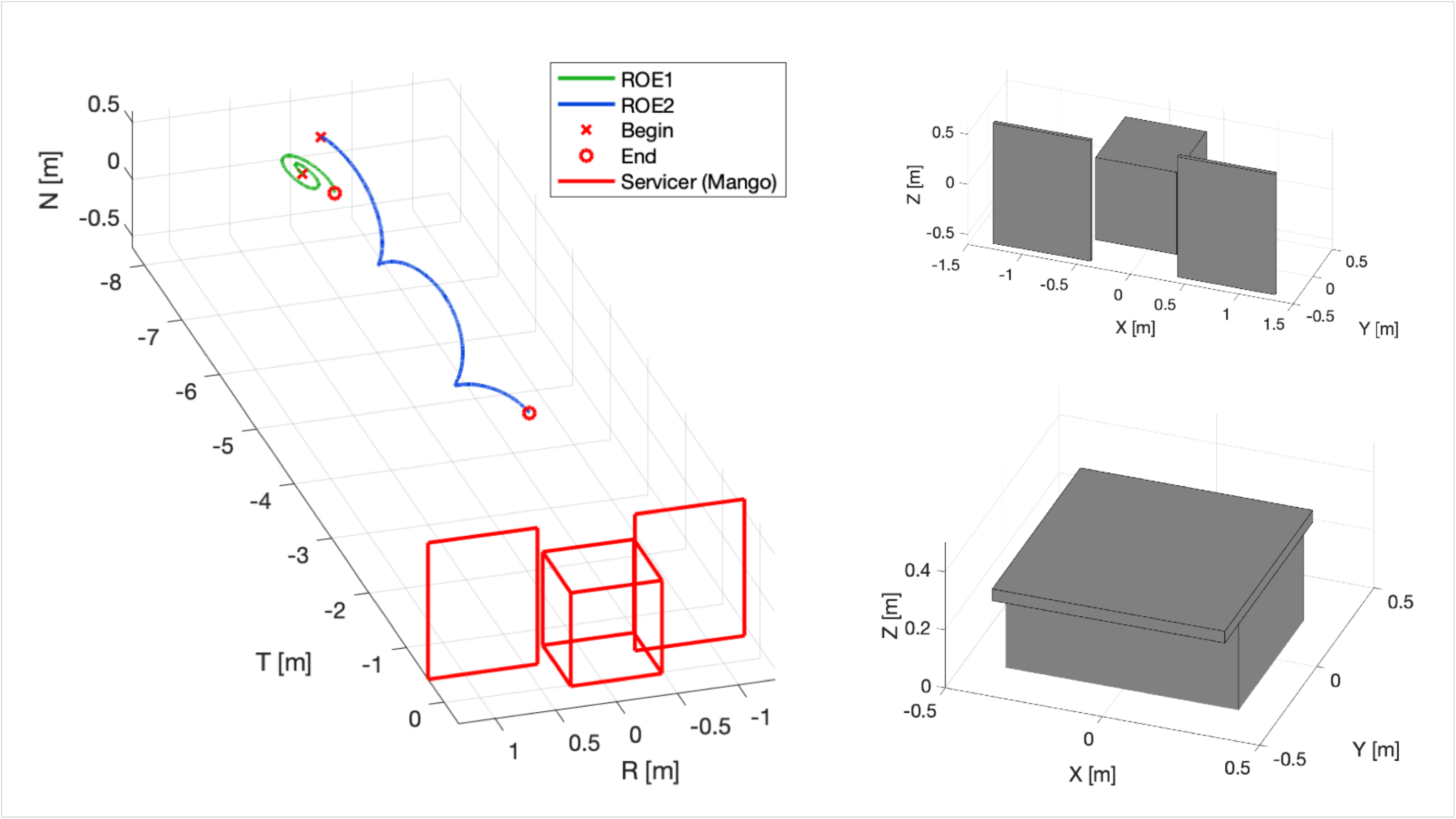}
	\caption{(\emph{Left}) Relative trajectories of the target (Tango) with respect to the servicer (Mango). (\emph{Right}) Simplified models of Mango (\emph{Top}) and Tango (\emph{Bottom}).}
	\label{fig:shirt_relative_trajectories}
\end{figure}

The orbital states of respective spacecraft are numerically propagated with a 1 second time step for two full orbits using the SLAB Satellite Software ($S^3$) \citep{Giralo2018GNSSTestbed}. Table \ref{tab:SHIRT simulation parameters} lists detailed simulation parameters which include rigorous force and torque models for ground-truth propagation. For model evaluation, the servicer and target spacecraft are modeled as Mango and Tango from the PRISMA mission \citep{PRISMA_chapter}. Spacecraft parameters for force and torque models are derived from \citet{Damico2010PHDThesis} and replicated in Table \ref{tab:mango_tango_parameters}. Note that the magnetic dipole moment of the servicer is set to zero, because that of the target is used to capture the differential perturbation between the two spacecraft arising from the Earth's magnetic field. To accurately propagate attitude motion, Mango and Tango are each modeled as an assembly of cuboid and rectangular plates as visualized in Fig.~\ref{fig:shirt_relative_trajectories}. The resulting relative trajectories of the target (Tango) with respect to the servicer (Mango) in the RTN frame are also visualized in Fig.~\ref{fig:shirt_relative_trajectories}. As expected from the initial ROE state, the target remains at approximately the same relative location with respect to the server in ROE1 throughout the simulation, whereas the servicer performs a spiral approach trajectory toward the target in ROE2.

\subsection{Image Acquisition}

\begin{figure}[!t]
	\centering
	\includegraphics[width=0.8\textwidth]{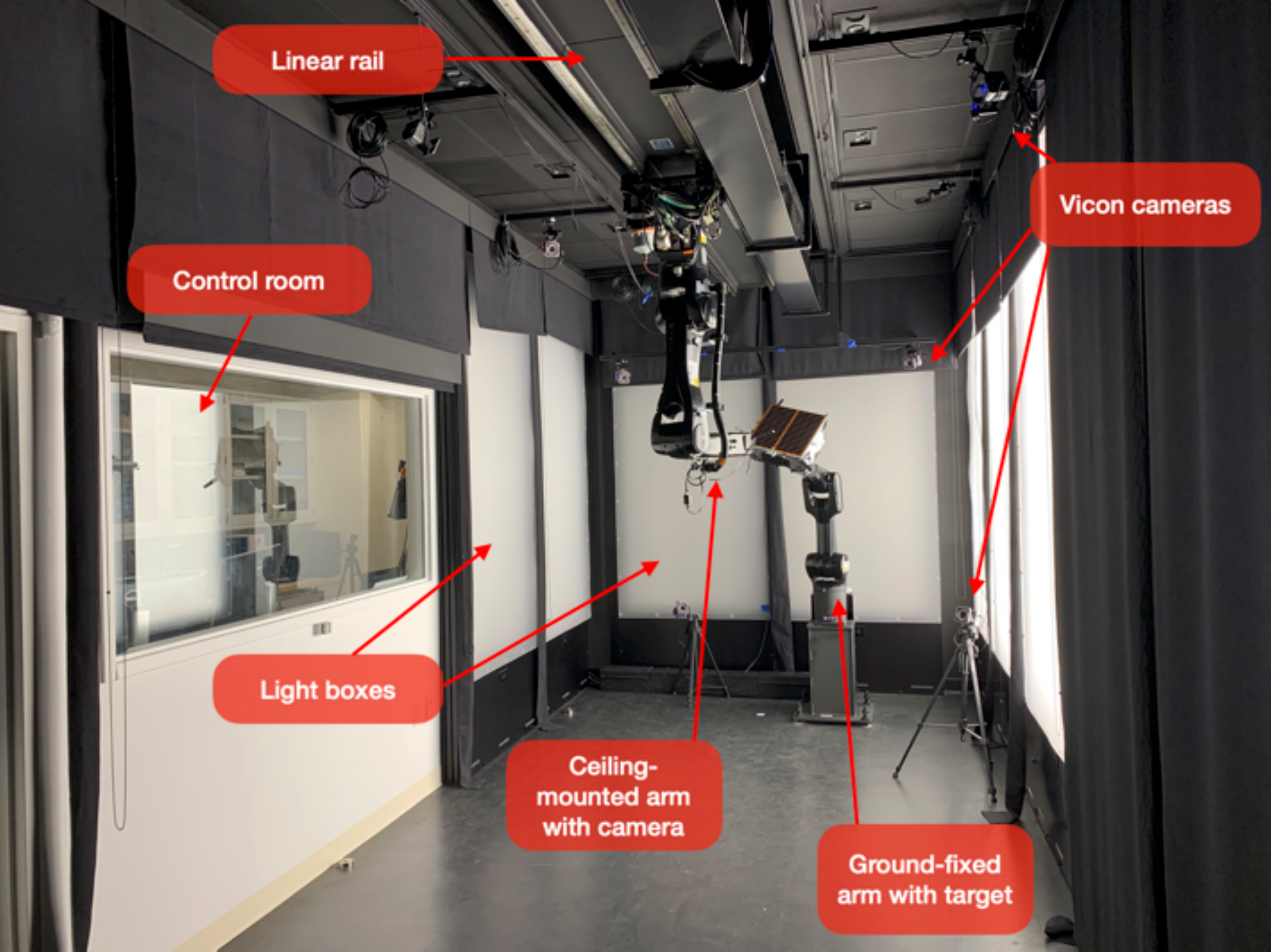}
	\caption{TRON facility at SLAB. Figure from \citet{Park2021AAS}.}
	\label{fig:TRON_overview}
\end{figure}

Once the relative trajectories are simulated, two sets of images are created for each rendezvous scenario with a capture interval of 5 seconds. The first set is the $\lightbox$ images captured with the robotic TRON facility \citep{Park2021AAS} at SLAB at Stanford University. As shown in Fig.~\ref{fig:TRON_overview}, the facility consists of two KUKA 6 DoF robot arms holding a camera and a half-scale mockup model of the Tango spacecraft, respectively. The facility provides the real-time pose of each robot's end-effector with respect to the global reference frame of the testbed; therefore, KUKA internal telemetry, along with the pose of infrared markers attached to both objects tracked by 12 Vicon Vero cameras \citep{Vicon}, can be jointly used to associate each image sample with high-accuracy pose labels. In particular, the facility is capable of reconstructing relative pose commands up to millimeter-level position accuracy and millidegree-level orientation accuracy upon calibration. TRON also includes 10 lightboxes \cite{LightBox} which are calibrated to emulate Earth albedo illumination in LEO. For more information on the facility, readers are referred to \citet{Park2021speedplus}.

\begin{figure}[!t]
	\centering
	\includegraphics[width=1.0\textwidth]{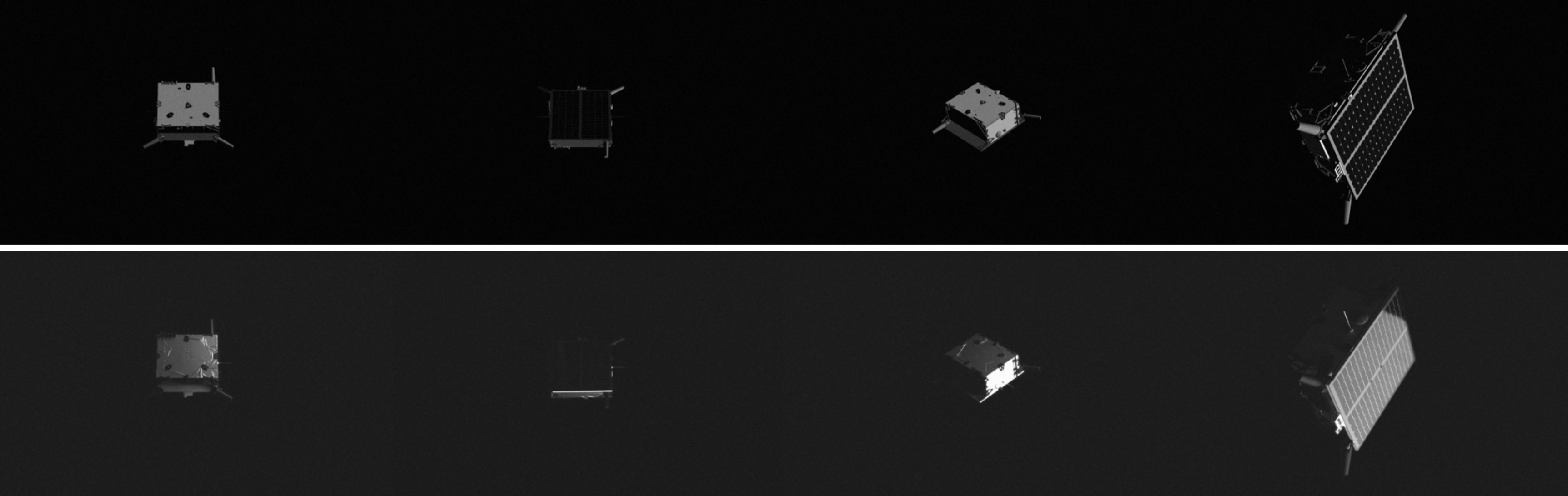}
	\caption{Samples of the $\synthetic$ (\emph{top}) and $\lightbox$ images (\emph{bottom}) with identical pose labels and matching illumination conditions.}
	\label{fig:samples_synthetic_lightbox}
\end{figure}

In order to simplify data acquisition, position of the target model is fixed within the facility, and the camera is always directed along the length of the room. Based on calibrated values for the offsets between each robot's end-effector and the object it holds \cite{Park2021AAS}, one can then convert the relative pose to be simulated into commands for the KUKA robot end-effectors. Each command is associated with a correct set of light boxes and proper light intensities to accurately simulate the desired albedo effect. Given that the reference trajectories are in a dawn-dusk sun-synchronous orbit, only 4 lightboxes located on opposite sides of the target model are used to emulate the effect of albedo light. Once captured, the $\lightbox$ images are processed via a procedure identical to that in SPEED+ \cite{Park2021speedplus}.

The second set of images is $\synthetic$ images rendered with the OpenGL-based Optical Stimulator (OS) \cite{SharmaBeierle2018_CNN, Beierle2019} software, using the intrinsic camera parameters estimated from the calibration of TRON. Unlike the SPEED+ $\synthetic$ images, the Earth is not inserted into the background since the camera is always pointed in the along-track direction in the reference trajectories. Comparisons of $\synthetic$ and $\lightbox$ images for identical pose labels are presented in Fig.~\ref{fig:samples_synthetic_lightbox}, which exhibits geometric consistency between images from both domains for identical pose labels. It also shows that the images captured from TRON successfully emulate the illumination conditions of their $\synthetic$ counterparts but with additional realism. Note that for CNN models trained on the SPEED+ $\synthetic$ images, the sequential images of the SHIRT $\synthetic$ domain do not present any domain gap since they originate from the same source as the training images. Therefore, the SHIRT $\synthetic$ images are intended to provide an indication of baseline performance, whereas its $\lightbox$ images are to be used for evaluation of filter performance across the domain gap.

In summary, SHIRT is a first-of-its-kind benchmark dataset comprising sequential images of the same target spacecraft with accurate pose labels. $\synthetic$ and $\lightbox$ images of the same trajectory in SHIRT exhibit consistency in rototranslational geometry and illumination conditions. Since CNN models are trained on $\synthetic$ images, the $\lightbox$ trajectory images of SHIRT allow for quantitative analyses of the performance of a CNN model and a navigation filter across the domain gap. Noting the scarcity of datasets featuring real images taken during rendezvous operations in space with fully annotated metadata, open-source datasets such as SHIRT and its future editions are invaluable for facilitating the validation efforts of vision-based GNC algorithms enabling proximity operations in space.
\section{Experiments} \label{section:experiments}

The proposed UKF with SPNv2 is tested on both $\synthetic$ and $\lightbox$ trajectories of SHIRT, but with more emphasis on the latter to examine the performance of the navigation filter across the domain gap. Specifically, the performance of SPNv2 only, the UKF with a constant process noise matrix $\bm{Q}_o$, and the UKF with ASNC are evaluated. Unless noted otherwise, the diagonal elements of $\bm{Q}_o$ are tuned so that $\bm{Q}_o = 1 \times 10^{-7}\bm{I}_{12 \times 12}$. When ASNC is activated, the sliding window length for covariance matching is set to $N_W = 60$, which corresponds to a 5 minute window for 5 seconds measurement intervals. The initial process noise covariance matrix is set to $\bm{Q}_o$ until $N_W$ images are observed, at which point covariance matching begins.

The UKF must also have estimates of the measurement noise covariance matrix $\bm{C}_\text{E}$ for the translation and rotation vector measurements from $\Ehead$ of SPNv2 (see Sec.~\ref{subsec:measurement model}). First, let $\bm{C}_\text{E}^\text{syn}$ denote the $6 \times 6$ covariance matrix derived from SPNv2's performance on the SPEED+ $\synthetic$ validation set \cite{park2022spnv2}. If the filter is tested on the $\synthetic$ trajectories of SHIRT, $\bm{C}_\text{E}$ can be simply set to $\bm{C}_\text{E}^\text{syn}$ since both SPEED+ and SHIRT $\synthetic$ images are created with OpenGL. On the other hand, if the filter is tested on the $\lightbox$ trajectories, $\bm{C}_\text{E}$ is instead set to $a\bm{C}_\text{E}^\text{syn}$, where $a$ is a positive scalar hyperparameter to adjust to the expected uncertainty of the vector measurements from a different domain. In this work, $a$ is set to 1000.

The filter state is initialized using the predictions of SPNv2 on the first image of the trajectory. Specifically, the relative angular velocity is computed based on the servicer's absolute measurement ($\bm{\omega}_{S,0}^S$) and assuming a non-tumbling target (i.e., $\bm{\omega}_T^T = \bm{0}_{3 \times 1}$). The target's initial relative velocity can be computed as $\hat{\bm{v}}_{T/S, 0} = \bm{\omega}_S^S \times \hat{\bm{t}}_{T/S, 0}$, where $\hat{\bm{t}}_{T/S, 0}$ is the target's position predicted from the first image with SPNv2. Then, ($\hat{\bm{t}}_{T/S, 0}, \hat{\bm{v}}_{T/S, 0}$) are converted to $\roe_0$ using the servicer's initial OE. The servicer's absolute orbital and attitude states are assumed to be known; however, the sensitivity of the filter's robustness with respect to absolute state errors is studied in Sec.~\ref{section:experiments:sensitivity-absolutestate}.

\subsection{Metrics}
For individual images, filter performance is evaluated using the translation error ($e_\text{t}$) and rotation error ($e_\text{q}$) defined in Eq.~\ref{eqn:pos rot metric individual}. Additionally, the pose error from SPEC2021 \cite{Park2021speedplus} is used as a singular metric wherever applicable and is given as
\begin{align}
	e_\textrm{pose} = e_\text{t} / \| \bm{t} \| + e_\textrm{q}
\end{align}
where $\bm{t}$ is the ground-truth translation vector of the sample, and $e_\text{q}$ is in radians. For batches of images, the mean translation, rotation, and pose errors are reported, respectively defined as
\begin{subequations} 
	\begin{align} 
		E_\text{t} &= \sum_{i=1}^N e_\text{t}^{(i)} \\
		E_\text{q} &= \sum_{i=1}^N e_\text{q}^{(i)} \\
		E_\text{pose} &= \sum_{i=1}^N e_\text{pose}^{(i)} \label{eqn:mean pose error}
	\end{align}
\end{subequations}
where $N$ is the number of images.

\subsection{SPNv2 Performance}

\begin{table*}[!t]
	\caption{Performance of SPNv2 on the SHIRT trajectories and SPEED+ $\synthetic$ validation and $\lightbox$ sets. Predictions from both Heatmap (H) and EfficientPose (E) heads are evaluated.}
	\label{tab:spnv2 on SHIRT}
	\centering
	\tabcolsep=0.1cm
	\begin{tabular}{@{}lllcc@{}}
		\toprule
		domain & trajectory & head & $E_\text{t}$ [m] & $E_\text{q}$ [$^\circ$] \\
		\midrule
		\multirow{4}{*}{$\synthetic$} & \multirow{2}{*}{ROE1} & H & 0.200 $\pm$ 0.119 & 2.360 $\pm$ 4.178 \\
        & & E & 0.115 $\pm$ 0.100 & 2.743 $\pm$ 4.472 \\
        \cmidrule(lr){2-5}
        & \multirow{2}{*}{ROE2} & H & 0.155 $\pm$ 0.098 & 1.289 $\pm$ 2.660 \\
        & & E & 0.061 $\pm$ 0.041 & 1.777 $\pm$ 2.811 \\
        \midrule
        \multirow{2}{*}{$\synthetic$} & \multirow{2}{*}{SPEED+} & H & - & 0.885 \\ & & E & 0.031 & - \\ 
        \midrule
        \multirow{4}{*}{$\lightbox$} & \multirow{2}{*}{ROE1} & H & 0.303 $\pm$ 0.349 & 17.168 $\pm$ 42.246 \\
        & & E & 0.175 $\pm$ 0.152 & 17.584 $\pm$ 41.854 \\
        \cmidrule(lr){2-5}
        & \multirow{2}{*}{ROE2} & H & 0.203 $\pm$ 0.231 & 4.669 $\pm$ 14.901 \\
        & & E & 0.100 $\pm$ 0.110 & 5.484 $\pm$ 16.418 \\
        \midrule
        \multirow{2}{*}{$\lightbox$} & \multirow{2}{*}{SPEED+} & H & - & 7.984 \\ & & E & 0.216 & - \\ 
		\bottomrule
	\end{tabular}
\end{table*}

The raw performance of SPNv2 alone is presented in Table \ref{tab:spnv2 on SHIRT} as a reference. The poses predicted by both $\Hhead$ and $\Ehead$ are evaluated on each SHIRT image without taking into account the sequential nature of these images. Recall from Section \ref{subsec:SPNv2} that $\Hhead$ outputs heatmaps associated with each keypoint of the target spacecraft which are converted to pose predictions via EP$n$P \citep{Lepetit2008EPnP}, whereas $\Ehead$ directly regresses the translation and rotation vectors. Table \ref{tab:spnv2 on SHIRT} also presents the reference performance of the particular configuration of SPNv2 used in this work on the SPEED+ $\lightbox$ test set \cite{Park2021speedplus}. Note that \citet{park2022spnv2} only reports the translation error from $\Ehead$ and rotation error from $\Hhead$ as these outputs have been shown to provide consistently more accurate predictions than their counterparts. This trend is similarly confirmed on the entire SHIRT dataset.

Table \ref{tab:spnv2 on SHIRT} shows that performance on the ROE2 $\lightbox$ images is better than on the SPEED+ $\lightbox$ images. Conversely, performance is much worse on the ROE1 $\lightbox$ domain, especially for rotation error, with near 17$^\circ$ mean error and extreme outliers as evidenced by the standard deviation of about 40$^\circ$. As explained in Sec.~\ref{section:shirt:trajsim}, the target has a varying inter-spacecraft distance and attitude throughout ROE2, which provides both easy and challenging images to SPNv2. On the other hand, the target is kept at a far range for the duration of ROE1, and target viewing angles are limited because it rotates about only one axis. Therefore, the images that SPNv2 receives are on the whole more challenging, which -- when combined with the domain gap -- leads to significantly degraded average performance on the ROE1 $\lightbox$ images.

\subsection{Filter Performance: $\synthetic$ Trajectories}

\begin{figure}[!t]
	\centering
	\begin{subfigure}{1.0\textwidth}
		\centering
		\includegraphics[width=\textwidth]{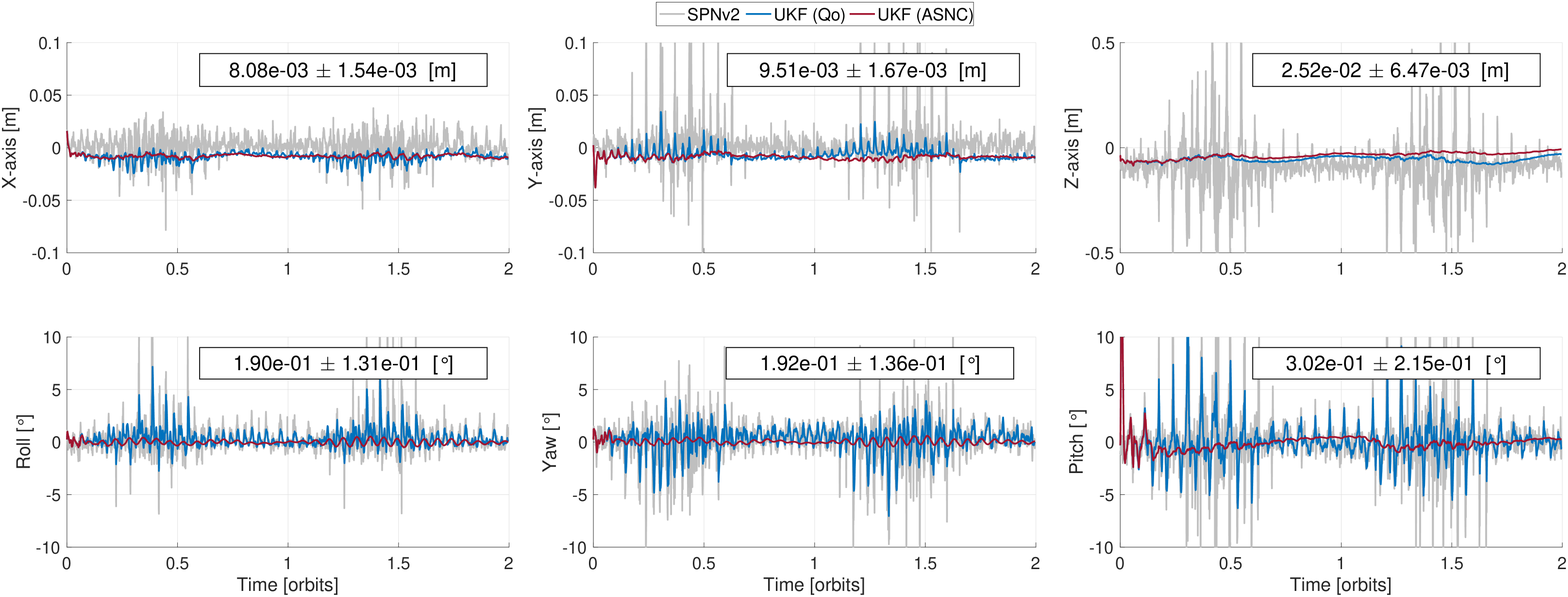}
		\caption{ROE1 $\synthetic$.}
		\label{fig:result:pose errors of synthetic:roe1}
	\end{subfigure}		
	\begin{subfigure}{1.0\textwidth}
		\centering
		\includegraphics[width=\textwidth]{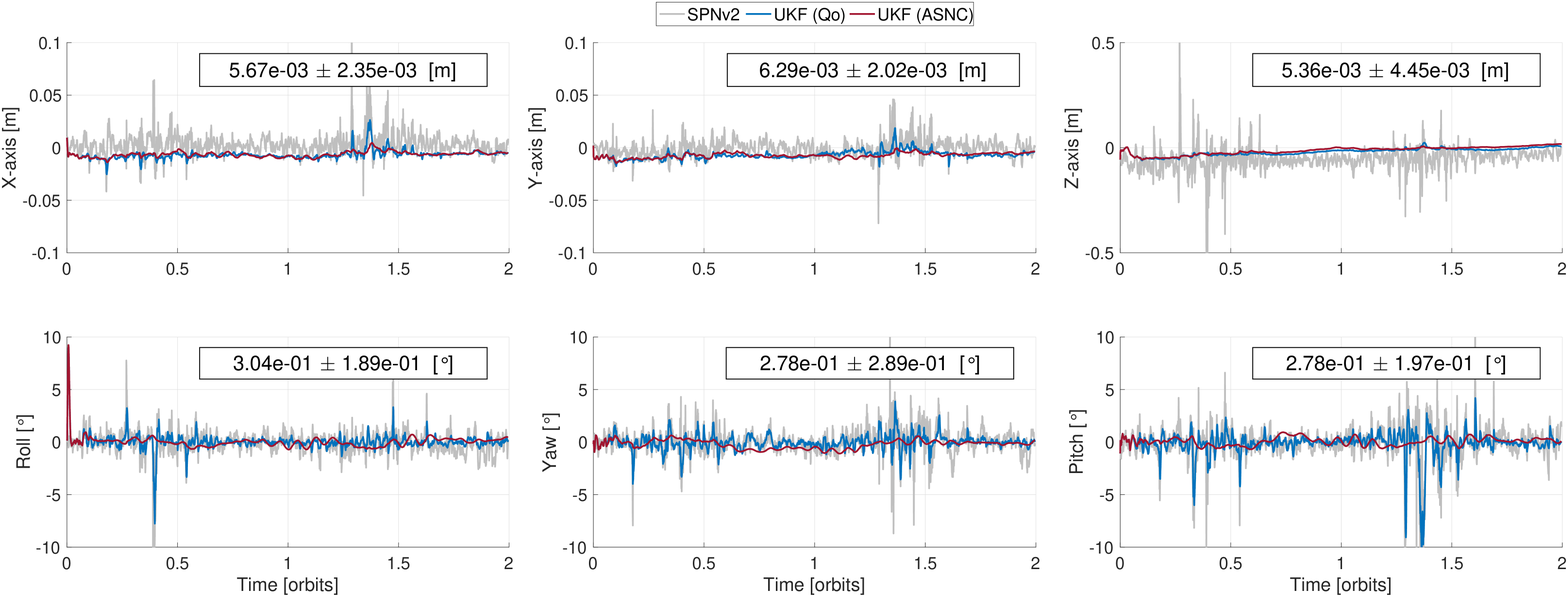}
		\caption{ROE2 $\synthetic$.}
		\label{fig:result:pose errors of synthetic:roe2}
	\end{subfigure}		
	\caption{Position and orientation errors of SPNv2 and different UKF configurations on the SHIRT $\synthetic$ trajectories. Position errors are given in the servicer's camera frame whose boresight is along $z$-axis. The boxed quantities denote the mean error and standard deviation of the UKF with ASNC during the second orbit.}
	\label{fig:result:pose errors of synthetic}
\end{figure}

The performance of the filter configurations with both constant $\bm{Q}_o$ and ASNC are first evaluated on the $\synthetic$ trajectories. Position and orientation errors of estimated states are plotted in Fig.~\ref{fig:result:pose errors of synthetic}. Pose estimates of SPNv2 alone are also plotted, for which position estimates are regressed from $\Ehead$ and orientation estimates are computed via P$n$P from keypoint measurements of $\Hhead$ as done in \citet{park2022spnv2}. It is observed that the integration of SPNv2 into the UKF reduces the position error throughout simulations. Furthermore, the activation of ASNC significantly smooths out the orientation error specifically. Note that even with ASNC, the filter initially struggles with the estimation of relative pitch error for ROE1. This is not unexpected, given that the target satellite only rotates about the servicer's pitch axis throughout the trajectory.

\subsection{Filter Performance: $\lightbox$ Trajectories}

\begin{figure}[!t]
	\centering
	\begin{subfigure}{1.0\textwidth}
		\centering
		\includegraphics[width=\textwidth]{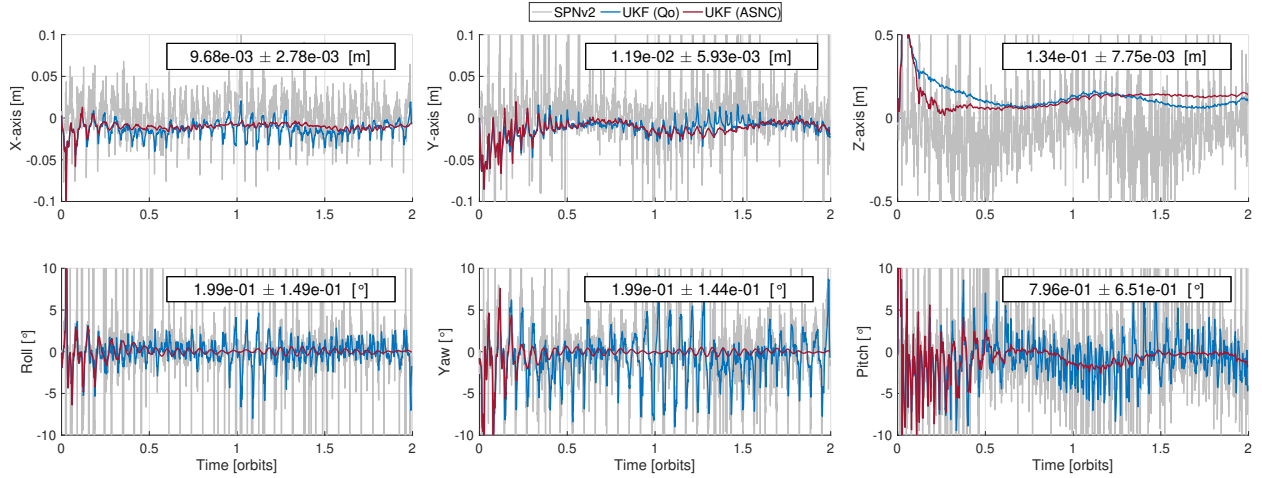}
		\caption{ROE1 $\lightbox$.}
		\label{fig:result:pose errors of lightbox:roe1}
	\end{subfigure}		
	\begin{subfigure}{1.0\textwidth}
		\centering
		\includegraphics[width=\textwidth]{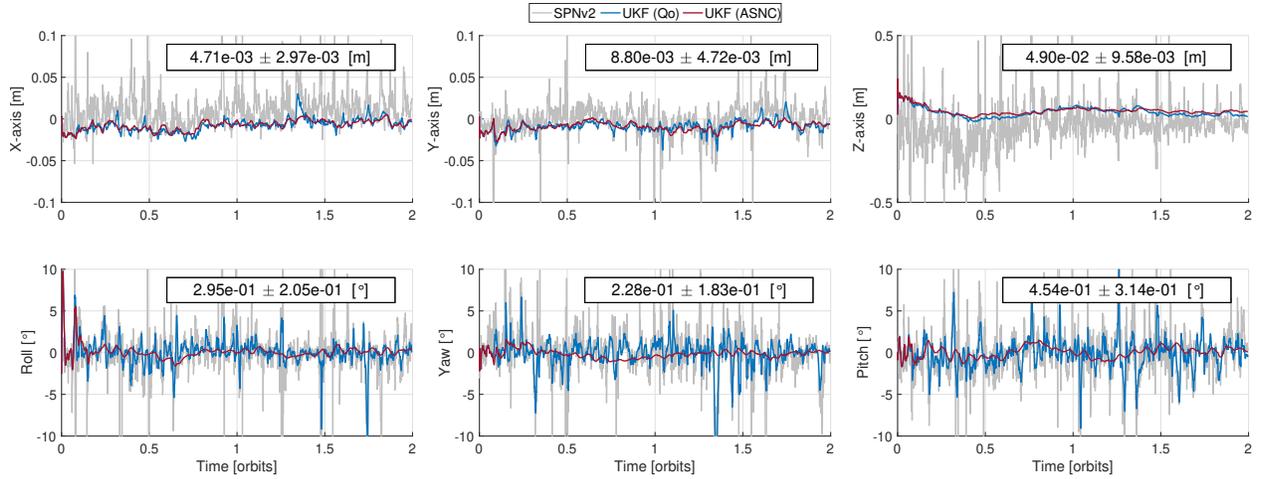}
		\caption{ROE2 $\lightbox$.}
		\label{fig:result:pose errors of lightbox:roe2}
	\end{subfigure}		
	\caption{Position and orientation errors of SPNv2 and different UKF configurations on the SHIRT $\lightbox$ trajectories. Position errors are given in the servicer's camera frame whose boresight is along the $z$-axis. The boxed quantities denote the mean error and standard deviation of the UKF with ASNC during the second orbit.}
	\label{fig:result:pose errors of lightbox}
\end{figure}

Next, the same set of experiments is performed on the more challenging SHIRT $\lightbox$ trajectories. Figure \ref{fig:result:pose errors of lightbox} shows the translation and orientation errors (Eq.~\ref{eqn:pos rot metric individual}) of SPNv2 alone and the UKF. It can be seen that when SPNv2 is used for pose prediction on the $\lightbox$ images without any filter integration, the predicted poses are much noisier than those on the $\synthetic$ trajectories due to the domain gap, as shown in Table \ref{tab:spnv2 on SHIRT} and visualized in Fig.~\ref{fig:result:pose errors of synthetic}. Note that the measurements are noisier for ROE1 than for ROE2 since its images are much more challenging for the reasons described earlier. On the other hand, when SPNv2 is integrated into the UKF with constant $\bm{Q}_o$, Figure \ref{fig:result:pose errors of lightbox} shows that the steady-state errors are significantly reduced for both position and orientation. Convergence behavior is further improved when ASNC is activated, as the estimated orientation in particular is smoothed out over the course of each trajectory, and error is kept below 2$^\circ$ at a steady state. Overall, Figure \ref{fig:result:pose errors of lightbox} indicates that it is possible to quickly converge to small steady-state errors on $\lightbox$ trajectory images when SPNv2, the UKF and ASNC are used in combination, even when the predictions of SPNv2 are noisy due to the domain gap.

\begin{figure}[!t]
	\centering
	\begin{subfigure}{\textwidth}
		\centering
		\includegraphics[width=\textwidth]{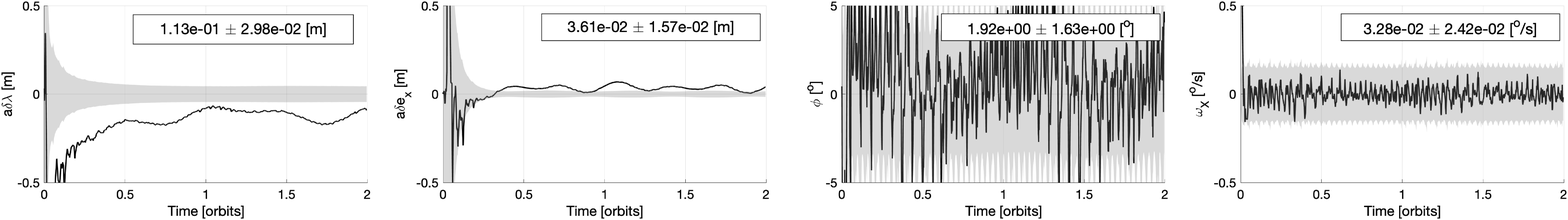}
		\includegraphics[width=\textwidth]{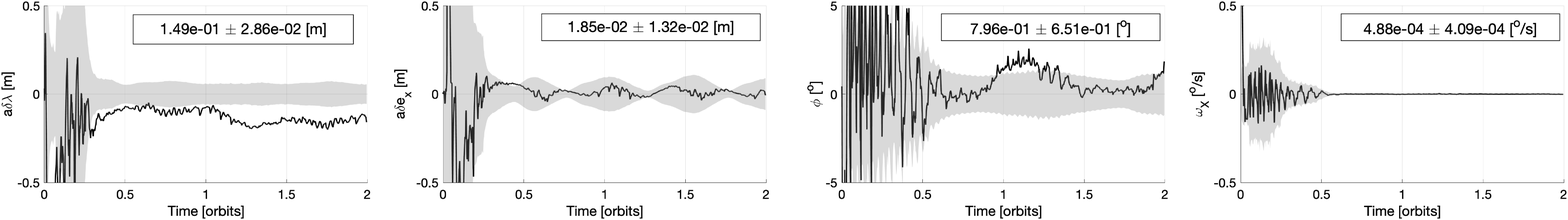}
		\caption{ROE1 $\lightbox$.}
	\end{subfigure}		
	\begin{subfigure}{\textwidth}
		\centering
		\includegraphics[width=\textwidth]{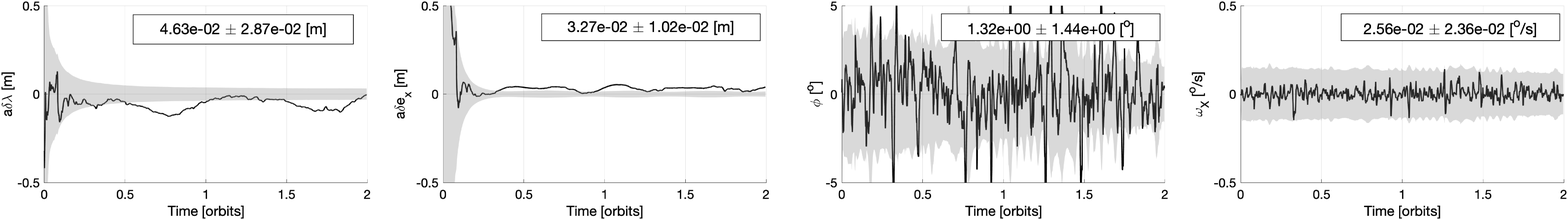}
		\includegraphics[width=\textwidth]{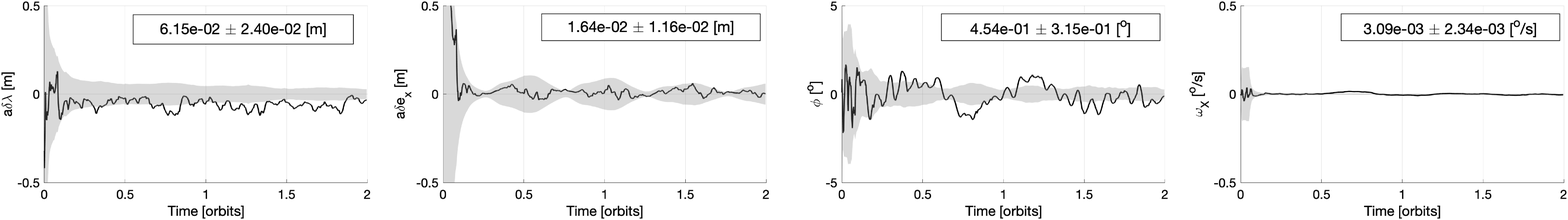}
		\caption{ROE2 $\lightbox$.}
	\end{subfigure}		
	\caption{Convergence and the associated formal 3-$\sigma$ bounds of the state vector elements on the $\lightbox$ trajectories. UKF with constant $\bm{Q}_o$ (\emph{top}) and ASNC (\emph{bottom}) are considered. The boxed quantities denote the mean error and standard deviation during the second orbit.}
	\label{fig:convergence}
\end{figure}

Next, the convergence behavior of a subset of the state vector is shown in Fig.~\ref{fig:convergence} for the ROE1 and ROE2 $\lightbox$ trajectories, respectively. Specifically, the relative longitude ($\delta\lambda$) and the $x$-component of the relative eccentricity vector ($\delta e_x$) (both scaled by the servicer's semi-major axis), the relative pitch angle ($\phi$), and the $x$-component of the relative angular velocity ($\omega_x$) are investigated. Observations from Fig.~\ref{fig:convergence} align with the results shown in Fig.~\ref{fig:result:pose errors of lightbox}. In particular, the estimated orientation and angular velocity component errors and their associated 3-$\sigma$ bounds are much lower and smoother with ASNC. Similarly, the 3-$\sigma$ bound for $\delta e_x$ better reflects the underlying uncertainty associated with the filter state estimate when ASNC is active. Overall, these results demonstrate that ASNC enables faster convergence, reduced steady-state errors, and state uncertainties which better reflect the uncertainties induced by adverse illumination conditions and measurement outliers due to the domain gap.

\subsection{\texorpdfstring{Sensitivity Analysis: $\bm{Q}_o$}{Q}}

\begin{figure}[!t]
	\centering
	\begin{subfigure}{\textwidth}
		\centering
		\includegraphics[width=0.45\textwidth]{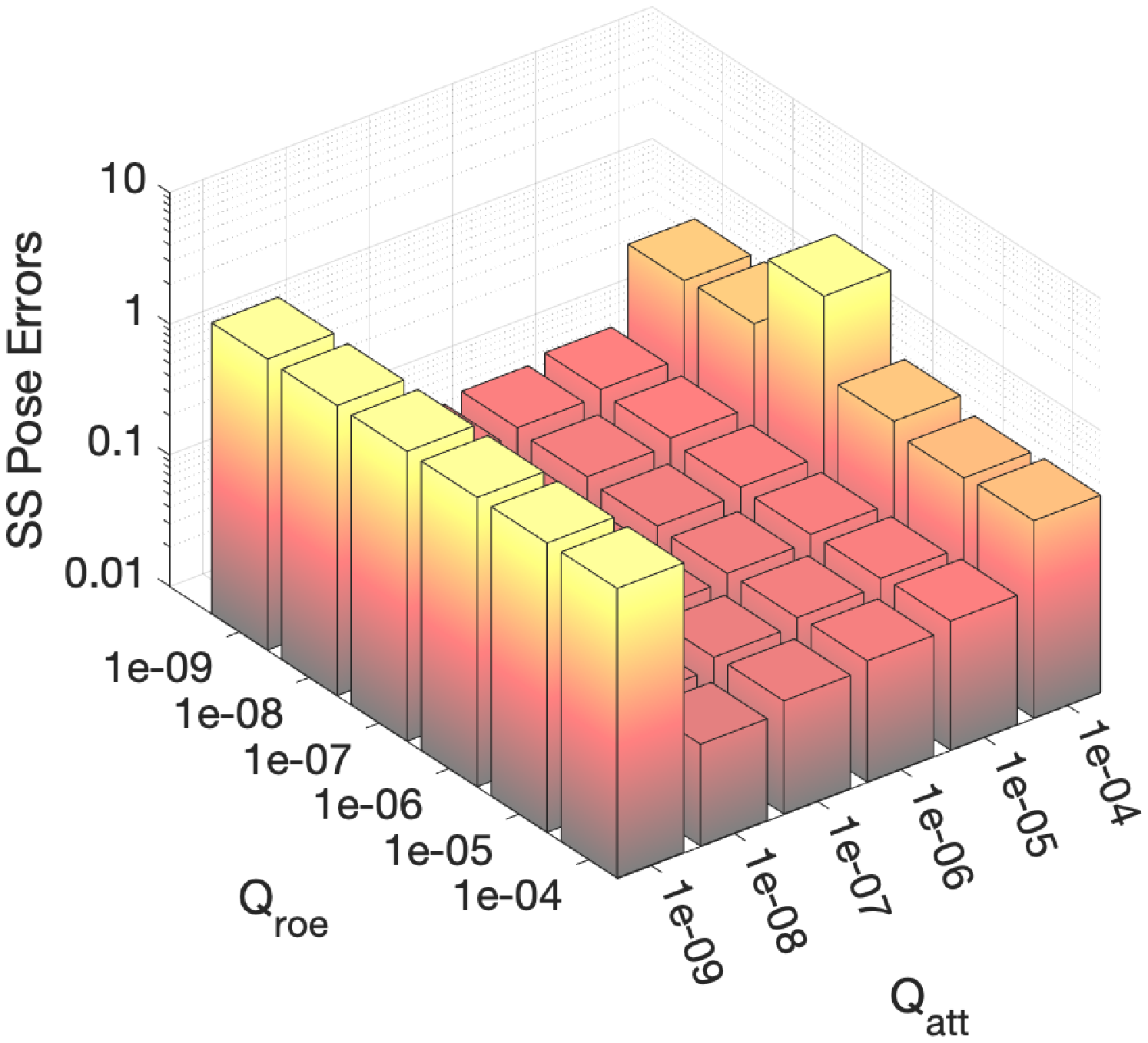}
		\includegraphics[width=0.45\textwidth]{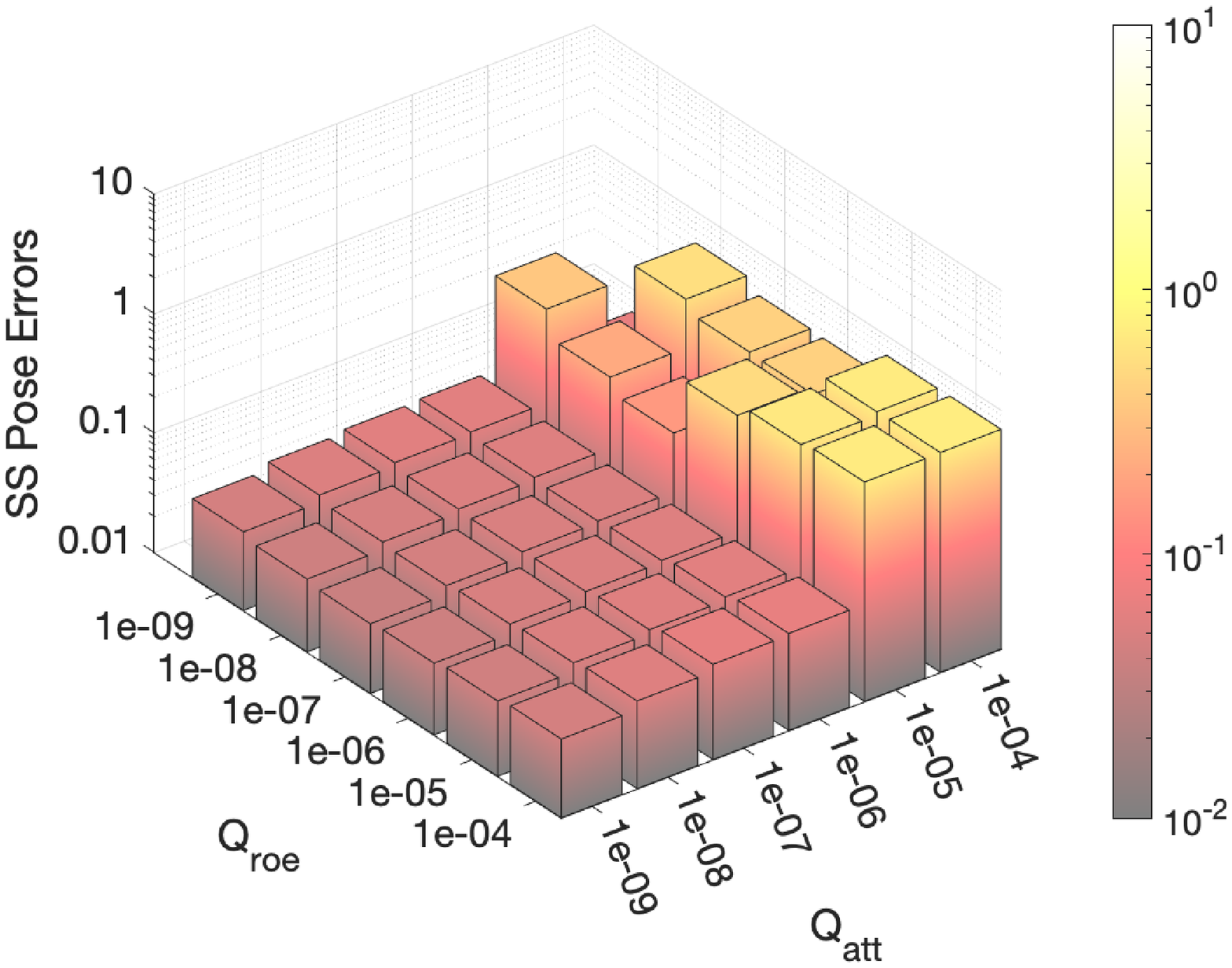}
		\caption{Constant $\bm{Q}_o$.}
		\label{fig:tuning Q:constant Q}
	\end{subfigure}		
	\begin{subfigure}{\textwidth}
		\centering
		\includegraphics[width=0.45\textwidth]{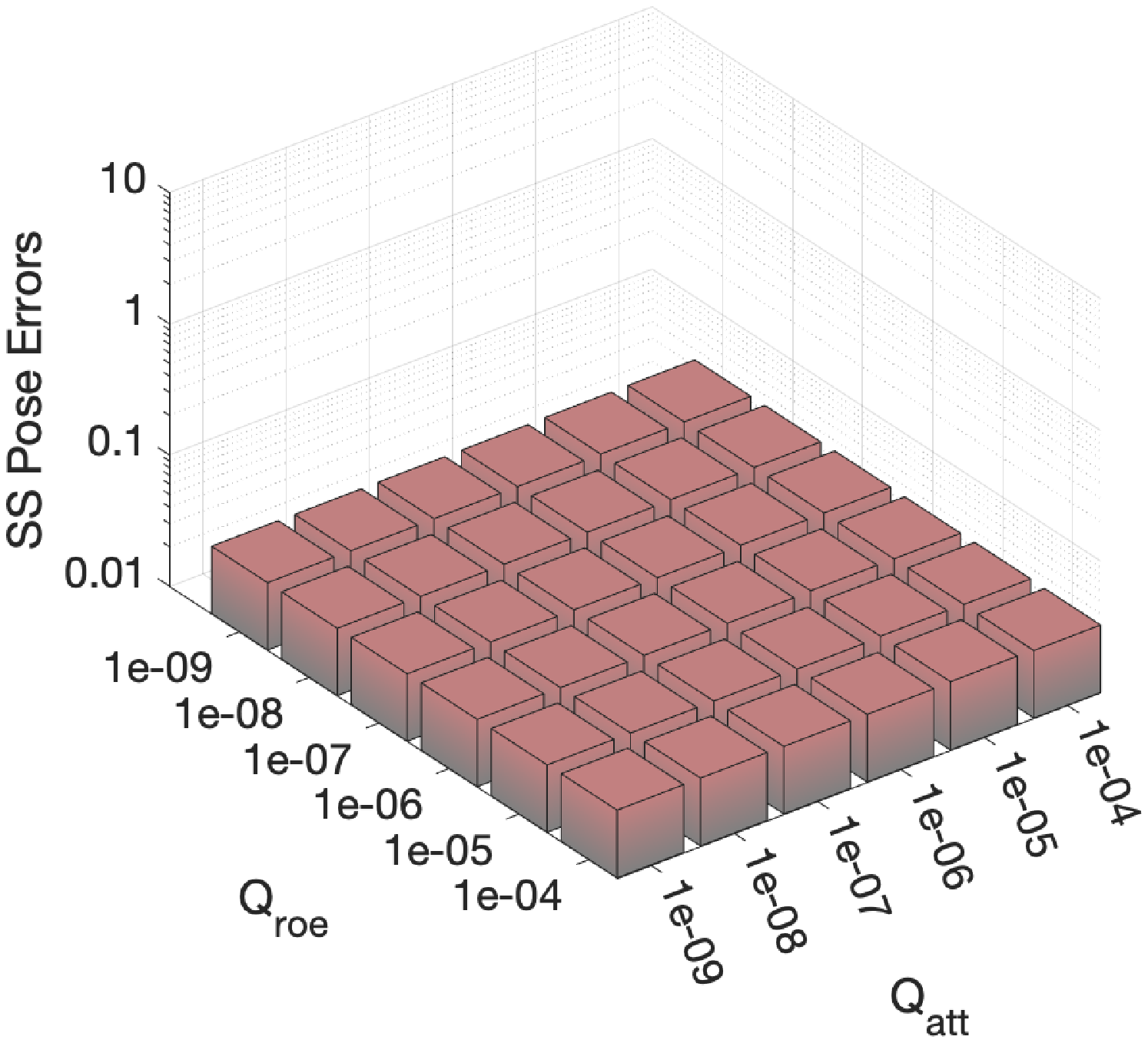}
		\includegraphics[width=0.45\textwidth]{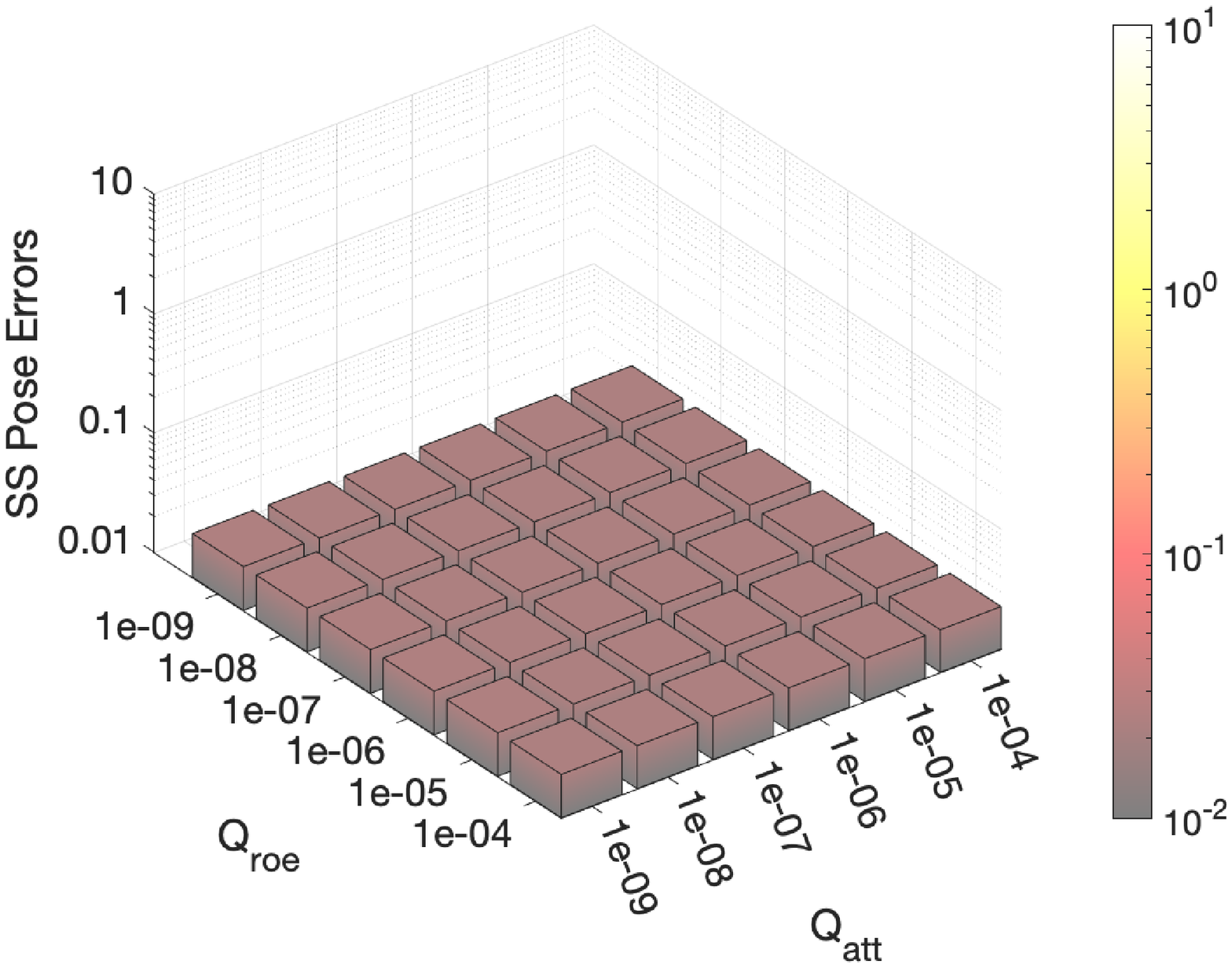}
		\caption{With ASNC.}
		\label{fig:tuning Q:ASNC}
	\end{subfigure}		
	\caption{The steady-state (SS) mean pose errors of the UKF in a logarithmic scale during the second orbit for the ROE1 (\emph{left}) and ROE2 (\emph{right}) $\lightbox$ trajectories when $Q_\text{roe}$ and $Q_\text{att}$ are varied.}
	\label{fig:tuning Q}
\end{figure}

The proposed UKF utilizes a matrix $\bm{Q}_o$ both as a constant process noise covariance matrix in the absence of ASNC and as its initializer when ASNC is activated. The advantage of ASNC is that it is free of meticulous process noise tuning due to its adaptive updates based on underlying continuous-time dynamics. However, manual tuning is still required during the initial filtering phase up until covariance matching can be performed. It is thus beneficial to conduct a sensitivity analysis into how different magnitudes of initial ASNC process noise may affect performance. In order to streamline the analysis, the process noise covariance matrix $\bm{Q}_o$ is modeled as
\begin{align}
	\bm{Q}_o = \begin{bmatrix} Q_\text{roe} \bm{I}_{6 \times 6} & \bm{0}_{6 \times 6} \\ \bm{0}_{6 \times 6} & Q_\text{att} \bm{I}_{6 \times 6} \end{bmatrix}
\end{align}
where the scalar parameters ($Q_\text{roe}, Q_\text{att}$) respectively tune the magnitudes of the uncertainties associated with orbital and attitude motion. These parameters are each varied from $1 \times 10^{-9}$ to $1 \times 10^{-4}$ in tenfold increments.

Figure \ref{fig:tuning Q} visualizes the steady-state mean pose errors during the second orbit (Eq.~\ref{eqn:mean pose error}) for each pair of values of $Q_\text{roe}$, $Q_\text{att}$. When constant $\bm{Q}_o$ is used, it is clear from Fig.~\ref{fig:tuning Q:constant Q} that convergence depends heavily on the magnitude of $\bm{Q}_o$. An interesting observation is that the steady-state error is affected largely by the magnitude of the process noise for relative attitude motion; regardless of the magnitude of $Q_\text{roe}$, the pose errors are exceptionally higher when $Q_\text{att}$ is either too high or too low. This is particularly true for ROE1 where the target rotates only about one axis. Since the observability of target attitude is severely limited, insufficient characterization of the uncertainties associated with relative attitude motion will significantly harm filter performance. In contrast, the target's attitude is visible from many different viewpoints in ROE2, so a small magnitude of $Q_\text{att}$ does not negatively affect filter convergence. Observe that when ASNC is activated, however, it is immediately obvious that for all rendezvous scenarios and cases, the filter converges to consistently small steady-state mean pose errors. Overall, Figure \ref{fig:tuning Q} shows that, as long as the initial process noise covariance matrix is set to a reasonable magnitude, ASNC ensures that the filter will converge to a steady state with low pose error.

\subsection{Sensitivity Analysis: Absolute State Noise} \label{section:experiments:sensitivity-absolutestate}

The aforementioned results are all obtained assuming perfect knowledge of the servicer's absolute state at every time step. In reality, the servicer's absolute state must also be estimated. In LEO, a satellite's absolute position can be estimated up to a decimeter-level using GNSS measurements \citep{Hauschild2021GNSSBroadcast} and its orientation up to an arcsecond-level using star trackers \citep{Liebe1995StarTrackers}. To assess the robustness of the UKF to noise in the servicer's absolute state estimates, 1000 Monte Carlo (MC) simulations are performed by injecting random noise characterized by Table \ref{tab:absolute state noise} into the servicer's true absolute state at each time step. Specifically, this work considers two cases: 1) a moderate case which reflects the nominal level of noise in LEO assuming the availability of GNSS receivers and star trackers on the servicer, and 2) a conservative case in which the noise levels are increased by the factor of 20 from the moderate case.

\begin{figure}[!t]
	\centering
 	\begin{subfigure}{\textwidth}
		\centering
	    \includegraphics[width=1.0\textwidth]{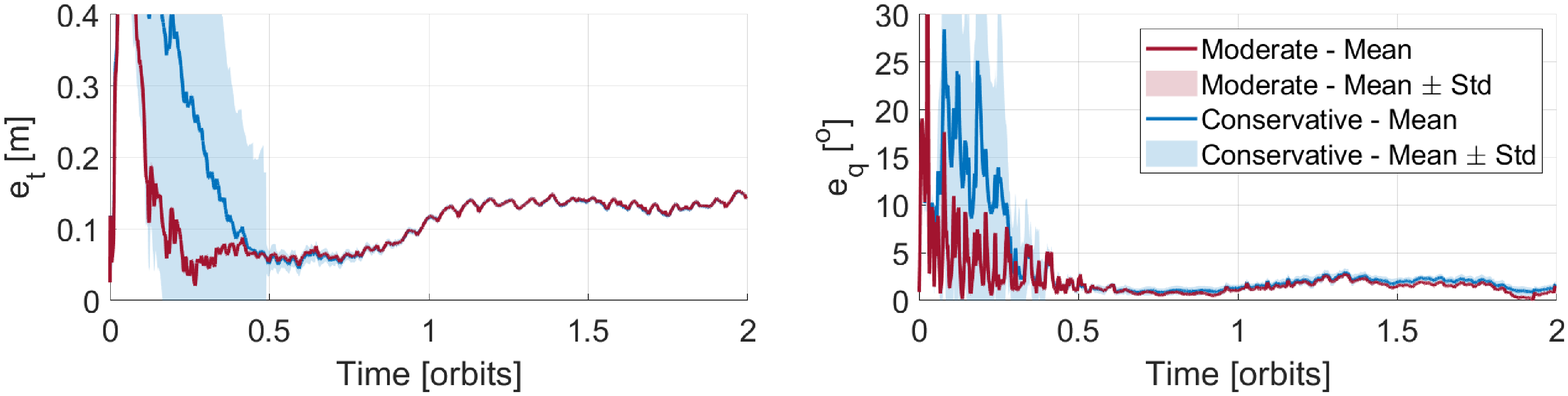}
        \caption{ROE1 $\lightbox$.}
    \end{subfigure}
  	\begin{subfigure}{\textwidth}
		\centering
	    \includegraphics[width=1.0\textwidth]{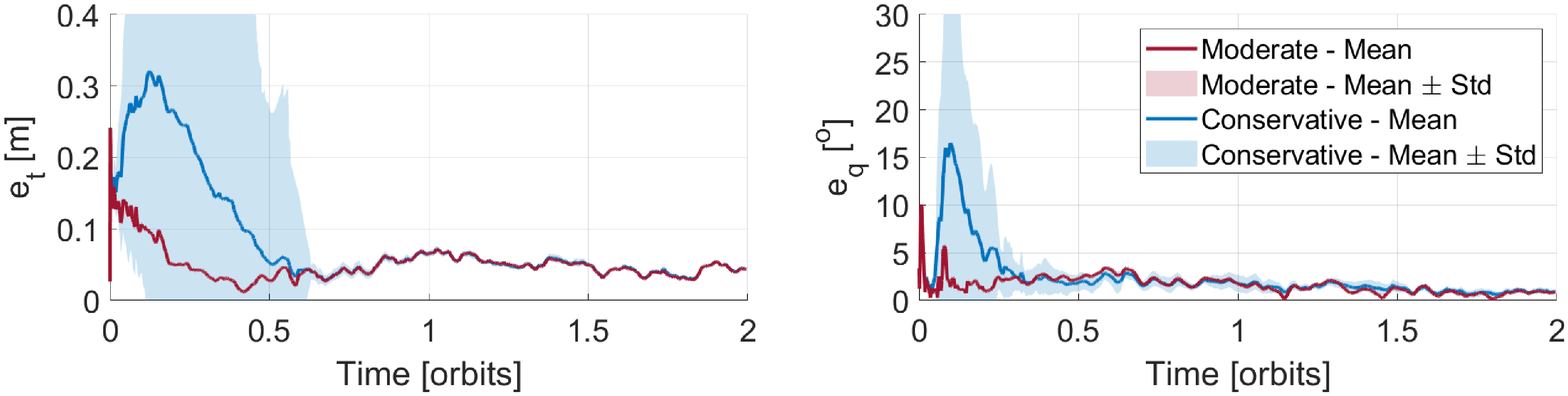}
        \caption{ROE2 $\lightbox$.}
    \end{subfigure}
	\caption{Monte Carlo simulations of the UKF with ASNC under perturbed absolute service orbit and attitude knowledge.}
	\label{fig:montecarlo}
\end{figure}

The results of MC simulations using the UKF with ASNC on the $\lightbox$ trajectories are shown in Fig.~\ref{fig:montecarlo}, which displays the mean and standard deviation of the estimated translation and orientation states. For the moderate case, all 1000 simulations converge to the same tracking patterns through the trajectories with extremely small deviations. In the conservative case, the CNN-powered UKF shows degraded performance during the first half-orbit of the convergence phase. However, it converges to small steady-state errors afterwards. The UKF with ASNC is therefore robust to varying quality of the servicer's absolute state knowledge.

\begin{table*}[!t]
	\caption{Standard deviation of the noise injected to the servicer's absolute state knowledge during MC simulations.}
	\label{tab:absolute state noise}
	\centering
	\tabcolsep=0.1cm
	\begin{tabular}{@{}lcccc@{}}
		\toprule
		Error Case & $\sigma_r$ [m] & $\sigma_v$ [cm/s] & $\sigma_q$ [arcsec] & $\sigma_w$ [arcsec/s] \\
		\midrule
		Moderate & 0.5 & 0.05 & 5 & 1 \\
        Conservative & 10 & 1 & 100 & 20 \\
		\bottomrule
	\end{tabular}
\end{table*}

\subsection{Comparison to Docking Requirements}

\begin{table*}[!t]
	\caption{Comparison of the UKF's steady-state errors against the design requirements of the MDS of the Orbital Express vehicles \mbox{\cite{Christiansen2008DockingSM}}.}
	\label{tab:mds requirement}
	\centering
	\tabcolsep=0.1cm
	\begin{tabular}{@{}lccc@{}}
		\toprule
		Parameter [unit] & Design Req. & CNN-UKF (ROE1) & CNN-UKF (ROE2) \\
		\midrule
		Axial capture distance [cm] & 15 & 13.42 $\pm$ 0.78 & 4.90 $\pm$ 0.96 \\
		Lateral misalignment [cm] & 5 & 1.63 $\pm$ 0.33 & 1.05 $\pm$ 0.45 \\
		Linear constant velocity [cm/s] & 3 & 0.0150 $\pm$ 0.0013 & 0.0061 $\pm$ 0.0010 \\
		Angular capture misalignment (pitch/yaw) [${}^\circ$] & 5 & 0.85 $\pm$ 0.62 & 0.56 $\pm$ 0.27 \\
		Angular capture misalignment (roll) [${}^\circ$] & 5 & 0.20 $\pm$ 0.15 & 0.30 $\pm$ 0.21 \\
		\bottomrule
	\end{tabular}
\end{table*}

Finally, to assess the filter's performance in the context of typical pose accuracy requirements imposed by rendezvous and docking processes, results are compared to the design requirements of the Mechanical Docking System (MDS) of the Orbital Express mission \cite{Christiansen2008DockingSM}. The comparison is justified by assuming that steady-state errors in the estimated relative pose during the close-proximity rendezvous would carry on to the ensuing docking process. In Table \ref{tab:mds requirement}, UKF ROE state estimates are converted to a Cartesian relative position and velocity, and position is broken into errors in lateral and axial components. Orientation error is converted into roll-pitch-yaw angles for comparison. Table \ref{tab:mds requirement} indicates that the relative orbital and attitude states estimated by the UKF and SPNv2 during the v-bar hold (ROE1) and approach (ROE2) trajectories fall well within the docking requirements posed for the OE mission. The lateral misalignment of the proposed UKF is on the centimeter level which is much less than the 5 cm requirement, and for all the other metrics apart from the axial capture distance, the UKF achieves steady-state errors at least an order of magnitude smaller than the requirements despite using a single low SWaP-C monocular camera.

Overall, the experimental results demonstrate that the integration of SPNv2 into the UKF and adaptive updates of the filter's process noise covariance enable a remarkable level of navigation performance in spaceborne proximity rendezvous scenarios. Performance is validated on both the $\synthetic$ and $\lightbox$ trajectories of SHIRT.

\section{Conclusion} \label{sec:conclusion}

This paper presents a complete navigation pipeline which combines the Spacecraft Pose Network v2 (SPNv2), a convolutional neural network for vision-based spacecraft pose estimation across the domain gap, with an adaptive Unscented Kalman Filter (UKF) to enable robust, accurate tracking of the position and orientation of a known noncooperative target spacecraft in close-range rendezvous scenarios. SPNv2 is trained exclusively on easily available synthetic images; therefore, to improve the convergence and accuracy of the filter across the domain gap, the process noise covariance matrix for relative orbit and relative attitude motion is adaptively updated at each time step via adaptive state noise compensation. As part of this process, a new analytical process noise model for relative attitude motion is derived and implemented. The paper also introduces the Satellite Hardware-in-the-loop Rendezvous Trajectories (SHIRT) dataset which consists of sequential images of two close-range rendezvous trajectories simulated with high-fidelity dynamics and kinematics models. Two image domains -- $\synthetic$ and $\lightbox$ -- are produced for each trajectory and present very different visual characteristics for the same target spacecraft. SHIRT therefore enables a comprehensive side-by-side comparison of a navigation filter's performance across a domain gap. The proposed UKF, which uses SPNv2 trained on synthetic images as an image processor, is shown to reach sub-decimeter-level position and degree-level orientation errors at steady-state on both domains of the SHIRT trajectories, successfully bridging the domain gap present in the dataset. Further analyses reveal that the proposed architecture is robust to different choices of magnitude for the initial process noise covariance matrix and varying noise levels in the absolute state knowledge of the servicer spacecraft.

The contributions made in this paper also identify limitations. One is the restricted range of inter-spacecraft separation simulated in SHIRT with a maximum distance of around 8 meters. This is due to the physical constraints of the robotic testbed used to create the SHIRT $\lightbox$ images. To enhance the utility of future image datasets, it is necessary to be able to simulate larger separations or docking sequences, which are another pivotal technical components required to support safe and autonomous servicing missions. Second, the SHIRT dataset simulates a slowly tumbling target spacecraft, which limits any robustness analyses of the proposed UKF design and ASNC with respect to target tumbling rate. Third, this work does not consider the on-board computational efficiency of the SPNv2 model during inference. Finally, this work assumes the knowledge of the target spacecraft's shape. This assumption must be relaxed if the target shape model is not available during the preliminary phases of an on-orbit servicing or debris removal mission.
\section{Appendix: Process Noise Covariance Models} 

\subsection{ROE State} \label{appendix:Process Noise Models ROE}
For $\roe^\prime = \bm{\alpha}_T - \bm{\alpha}_S$, where $\bm{\alpha}$ is a vector of equinoctial elements, the linear mapping matrix $\bm{X}_{k,i}^\prime$ for $i \in \{r, t, n\}$ is given as \cite{Stacey2022CovModel}
\begin{subequations}
\begin{align}
	\bm{X}_{k}^{r\prime} &= \Delta t_k \bm{\Gamma}_{k}^r \bm{\Gamma}_{k}^{r\top} + \frac{3n\Delta t_k^2}{4a} \begin{bmatrix} \bm{0}_{5 \times 5} & \bm{S}_r \\ \bm{S}_r^\top & \frac{n}{a} \bar{A}^2 \Delta t_k - 2\bar{A}\bar{K} \end{bmatrix}, \\
	\bm{X}_{k}^{t\prime} &= \Delta t_k \	\bm{\Gamma}_{k}^t \bm{\Gamma}_{k}^{t\top} + \frac{3n\Delta t_k^2}{4a} \begin{bmatrix} \bm{0}_{5 \times 5} & \bm{S}_t \\ \bm{S}_t^\top & \frac{n}{a} \bar{B}^2 \Delta t_k - 2\bar{B}\bar{L} \end{bmatrix}, \\
	\bm{X}_{k}^{n\prime} &= \Delta t_k 	\bm{\Gamma}_{k}^n \bm{\Gamma}_{k}^{n\top}.
\end{align}
\end{subequations}
Here, $\bm{\Gamma}_k = [\bm{\Gamma}_k^r ~~~\bm{\Gamma}_k^t ~~~ \bm{\Gamma}_k^n ] \in \mathbb{R}^{6 \times 3}$ denotes the time derivative of the equinoctial elements given by the Gauss Variational Equations,
\begin{align} \label{eqn:Gamma from GVE}
	\bm{\Gamma}_k = \begin{bmatrix} \bar{A} & \bar{B} & 0 \\ \bar{C} & \bar{D} & \bar{E} \\ \bar{F} & \bar{G} & \bar{H} \\ 0 & 0 & \bar{I} \\ 0 & 0 & \bar{J} \\ \bar{K} & \bar{L} & \bar{M} \end{bmatrix},
\end{align}
where the elements of $\bm{\Gamma}_k$ are based on the servier's OE ($\bm{\alpha}_S$) at $t_k$, and $\bm{S}_r = -[ \bar{A}^2 ~~ \bar{A}\bar{C} ~~ \bar{A}\bar{F} ~~ 0 ~~ 0]^\top$, $\bm{S}_t = -[ \bar{B}^2 ~~ \bar{B}\bar{D} ~~ \bar{B}\bar{G} ~~ 0 ~~ 0]^\top$. The barred elements are available in Stacey and D'Amico \cite{Stacey2022CovModel} and are not reproduced here for brevity.

\subsection{Attitude State} \label{appendix:Process Noise Models Attitude}

The sub-matrices of Eq.~\ref{eqn:X matrix for attitude in A, B, C} are given as
\begin{subequations}
\begin{multline}
	\bar{\bm{A}}_i = \int_{t_{k-1}}^{t_k} \bm{\Lambda}_{1,i}(t_k - \tau) \bm{\Lambda}_{1,i}(t_k - \tau)^\top d\tau = \frac{\Delta t_k^3}{3} \bm{e}_i \bm{e}_i^\top + \zeta_{c_1 c_1} \bm{W}_{1,i} \bm{W}_{1,i}^\top + \zeta_{s_1 s_1} \bm{V}_{1,i} \bm{V}_{1,i}^\top \\ 
	+ \zeta_{tc_1} (\bm{e}_i \bm{W}_{1,i}^\top + \bm{W}_{1,i} \bm{e}_i^\top) + \zeta_{ts_1} (\bm{e}_i \bm{V}_{1,i}^\top + \bm{V}_{1,i} \bm{e}_i^\top) + \zeta_{c_1 s_1} (\bm{W}_{1,i} \bm{V}_{1,i}^\top + \bm{V}_{1,i} \bm{W}_{1,i}^\top), 
\end{multline}
\begin{multline}
	\bar{\bm{B}}_i = \int_{t_{k-1}}^{t_k} \bm{\Lambda}_{1,i}(t_k - \tau) \bm{\Lambda}_{2,i}(t_k - \tau)^\top d\tau = -\bigg( \frac{\Delta t_k^2}{2} \bm{e}_i \bm{e}_i^\top + \zeta_{ts_2}\bm{e}_i \bm{W}_{2,i}^\top + \zeta_{tc_2}\bm{e}_i\bm{V}_{2,i}^\top + \zeta_{c_1}\bm{W}_{1,i}\bm{e}_i^\top \\
	+ \zeta_{c_1s_2} \bm{W}_{1,i} \bm{W}_{2,i}^\top +  \zeta_{c_1c_2} \bm{W}_{1,i} \bm{V}_{2,i}^\top + \zeta_{s_1} \bm{V}_{1,i} \bm{e}_i^\top + \zeta_{s_1s_2}\bm{V}_{1,i}\bm{W}_{2,i}^\top + \zeta_{s_1c_2}\bm{V}_{1,i}\bm{V}_{2,i}^\top \bigg),
\end{multline}
\begin{multline}
	\bar{\bm{C}}_i = \int_{t_{k-1}}^{t_k} \bm{\Lambda}_{2,i}(t_k - \tau) \bm{\Lambda}_{2,i}(t_k - \tau)^\top d\tau = \Delta t_k \bm{e}_i \bm{e}_i^\top  + \zeta_{c_2 c_2} \bm{V}_{2,i} \bm{V}_{2,i}^\top + \zeta_{s_2 s_2} \bm{W}_{2,i} \bm{W}_{2,i}^\top \\
	+ \zeta_{c_2} (\bm{e}_i \bm{V}_{2,i}^\top + \bm{V}_{2,i} \bm{e}_i^\top) + \zeta_{s_2} (\bm{e}_i \bm{W}_{2,i}^\top + \bm{W}_{2,i} \bm{e}_i^\top) + \zeta_{c_2s_2} (\bm{W}_{2,i} \bm{V}_{2,i}^\top + \bm{V}_{2,i} \bm{W}_{2,i}^\top),
\end{multline}
\end{subequations}
where $\bm{W}_j = [\bm{\hat{\omega}}_j]_\times = [\bm{W}_{j,x} ~~ \bm{W}_{j,y} ~~ \bm{W}_{j,z}]$, $\bm{V}_j = [\bm{\hat{\omega}}_j]_\times^2 = [\bm{V}_{j,x} ~~ \bm{V}_{j,y} ~~ \bm{V}_{j,z}]$. Recall that $\bm{\omega}_1$ denotes $\bm{\omega}_{S/T, k}^T$, and $\bm{\omega}_2$ denotes $\bm{R}_{T/S, k}\bm{\omega}_{S, k}^S$. The $\zeta$ coefficients are then expressed analytically by evaluating the integrals of each term. Defining $c_1 = \cos\omega_1\Delta t_k$, $s_1 = \sin\omega_1\Delta t_k$, $c_2 = \cos \omega_2 \Delta t_k$, $s_2 = \sin \omega_2 \Delta t_k$, the coefficients are given as
\begin{subequations}
    \allowdisplaybreaks
	\begin{flalign*}
	\zeta_{c_1} &= \int_{t_{k-1}}^{t_k} \frac{1}{\omega_1}\big[ 1 - \cos \omega_1(t_k - \tau) \big] d\tau = \frac{1}{\omega_1}\bigg(\Delta t_k - \frac{s_1}{\omega_1} \bigg) &\\
	\zeta_{s_1} &= \int_{t_{k-1}}^{t_k} (t_k - \tau) - \frac{1}{\omega_1} \sin \omega_1(t_k - \tau) d\tau = \frac{\Delta t_k^2}{2} + \frac{c_1 - 1}{\omega_1^2}  &\\
	\zeta_{tc_1} &=  \int_{t_{k-1}}^{t_k} \frac{t_k - \tau}{\omega_1}\big[ 1 - \cos \omega_1(t_k - \tau) \big] d\tau = \frac{1}{\omega_1} \bigg[ \frac{\Delta t_k^2}{2} + \frac{1 - c_1 - \Delta t_k \omega_1 s_1}{\omega_1^2} \bigg] &\\
	\zeta_{ts_1} &=  \int_{t_{k-1}}^{t_k} (t_k - \tau) \bigg[ (t_k - \tau) - \frac{1}{\omega_1} \sin \omega_1(t_k - \tau) \bigg] d\tau = \frac{\Delta t_k^3}{3} + \frac{1}{\omega_1^2} \bigg(\Delta t_k c_1 - \frac{s_1}{\omega_1} \bigg) &\\
	\zeta_{c_1c_1} &= \int_{t_{k-1}}^{t_k} \frac{1}{\omega^2} \big[ 1 - \cos \omega_1(t_k - \tau) \big]^2 d\tau = \frac{1}{\omega_1^2} \bigg[\frac{3}{2}\Delta t_k + \frac{s_1}{2\omega_1}\bigg( c_1 - 4 \bigg) \bigg]&\\
	\zeta_{s_1s_1} &= \int_{t_{k-1}}^{t_k} \bigg[(t_k - \tau) - \frac{1}{\omega_1} \sin \omega_1(t_k - \tau)\bigg]^2 d\tau = \frac{\Delta t_k^3}{3} + \frac{1}{\omega_1^2} \bigg(\frac{\Delta t_k}{2} + 2\Delta t_k c_1 - \frac{2s_1}{\omega_1} - \frac{s_1c_1}{2\omega_1} \bigg)  &\\
	\zeta_{c_1s_1} &=  \int_{t_{k-1}}^{t_k} \frac{1}{\omega_1}\big[ 1 - \cos \omega_1(t_k - \tau) \big] \bigg[ (t_k - \tau) - \frac{1}{\omega_1} \sin \omega_1(t_k - \tau) \bigg] d\tau = \frac{(s_1 - \omega_1\Delta t_k)^2}{2\omega_1^3} &\\
	\zeta_{c_2} &= \int_{t_{k-1}}^{t_k} \big[1 - \cos \omega_2(t_k - \tau) \big] d\tau = \Delta t_k - \frac{s_2}{\omega_2} &\\
	\zeta_{s_2} &= \int_{t_{k-1}}^{t_k} -\sin \omega_2(t_k - \tau) d\tau = \frac{c_2 - 1}{\omega_2}  &\\
	\zeta_{tc_2} &= \int_{t_{k-1}}^{t_k} (t_k - \tau)\big[ 1 - \cos \omega_2(t_k - \tau) \big] d\tau = \frac{\Delta t_k^2}{2} + \frac{1 - c_2 - \Delta t_k \omega_2 s_2}{\omega_2^2}  &\\
	\zeta_{ts_2} &= \int_{t_{k-1}}^{t_k} - (t_k - \tau)\sin \omega_2(t_k - \tau) d\tau  = \frac{1}{\omega_2} \bigg( \Delta t_k c_2 - \frac{s_2}{\omega_2} \bigg)  &\\
	\zeta_{c_2c_2} &= \int_{t_{k-1}}^{t_k} \big[ 1 - \cos \omega_2(t_k - \tau) \big]^2 d\tau = \frac{3}{2}\Delta t_k + \frac{s_2}{2\omega_2}\bigg( c_2 - 4 \bigg) &\\
	\zeta_{s_2s_2} &= \int_{t_{k-1}}^{t_k}  \sin^2 \omega_2(t_k - \tau) d\tau = \frac{\Delta t_k}{2} - \frac{s_2c_2}{2\omega_2} &\\
	\zeta_{c_2s_2} &= \int_{t_{k-1}}^{t_k}  -\big[ 1 - \cos \omega_2(t_k - \tau)\big] \sin \omega_2(t_k - \tau) d\tau = -\frac{1}{2\omega_2} (1 - c_2)^2 &\\
	\zeta_{c_1c_2} &= \int_{t_{k-1}}^{t_k}  \frac{1}{\omega_1}\big[ 1 - \cos \omega_1(t_k - \tau) \big] (1 - \cos \omega_2(t_k - \tau)) d\tau  = \\
	&\qquad\qquad\qquad\qquad\qquad\qquad\qquad\qquad \frac{1}{\omega_1} \bigg( -\Delta t_k + \omega_1\zeta_{c_1} + \zeta_{c_2} + \frac{\omega_1s_1c_2 - \omega_2c_1s_2}{\omega_1^2 - \omega_2^2} \bigg) &\\
	\zeta_{c_1s_2} &= \int_{t_{k-1}}^{t_k}  \frac{1}{\omega_1}\big[ 1 - \cos \omega_1(t_k - \tau) \big] (-\sin \omega_2(t_k - \tau)) d\tau = \frac{1}{\omega_1} \bigg(\zeta_{s_2} + \frac{\omega_1s_1s_2 + \omega_2c_1c_2 - \omega_2}{\omega_1^2 - \omega_2^2} \bigg) &\\
	\zeta_{s_1c_2} &= \int_{t_{k-1}}^{t_k} \bigg[(t_k - \tau) - \frac{1}{\omega_1} \sin \omega_1(t_k - \tau)\bigg]  (1 - \cos \omega_2(t_k - \tau)) d\tau = \\
	&\qquad\qquad\qquad\qquad\qquad\qquad\qquad\qquad \frac{-\Delta t_k^2}{2} + \zeta_{s_1} + \zeta_{tc_2} + \frac{1}{\omega_1}\bigg(  \frac{\omega_2s_1s_2 + \omega_1c_1c_2 - \omega_1}{\omega_2^2 - \omega_1^2} \bigg) &\\
	\zeta_{s_1s_2} &= \int_{t_{k-1}}^{t_k} \bigg[(t_k - \tau) - \frac{1}{\omega_1} \sin \omega_1(t_k - \tau)\bigg] (-\sin \omega_2(t_k - \tau)) d\tau = \zeta_{ts_2} + \frac{1}{\omega_1} \bigg( \frac{\omega_2s_1c_2 - \omega_1c_1s_2}{\omega_1^2 - \omega_2^2} \bigg)
	\end{flalign*}
\end{subequations}

\section*{Funding Sources}
This work is partially supported by Taqnia International through contract 1232617-1-GWNDV.

\section*{Acknowledgments}
The authors would like to thank OHB Sweden for the 3D model of the Tango spacecraft used to create the images used in this article. The authors also thank Dr.~Nathan Stacey at Stanford's Space Rendezvous Laboratory for helpful discussions.

\bibliography{reference}

\begin{thebibliography}{59}
\newcommand{\enquote}[1]{``#1''}
\providecommand{\natexlab}[1]{#1}
\providecommand{\url}[1]{\texttt{#1}}
\providecommand{\urlprefix}{URL }
\expandafter\ifx\csname urlstyle\endcsname\relax
  \providecommand{\doi}[1]{\discretionary{}{}{}https://doi.org/#1}\else
  \providecommand{\doi}[1]{\discretionary{}{}{}\urlstyle{rm}\url{https://doi.org/#1}}\fi

\bibitem[{Reed et~al.(2016)Reed, Smith, Naasz, Pellegrino, and
  Bacon}]{Reed2016RestoreL}
Reed, B.~B., Smith, R.~C., Naasz, B.~J., Pellegrino, J.~F., and Bacon, C.~E.,
  \enquote{The {Restore-L} Servicing Mission,} \emph{AIAA Space 2016}, 2016.
\newblock \doi{10.2514/6.2016-5478}.

\bibitem[{Forshaw et~al.(2016)Forshaw, Aglietti, Navarathinam, Kadhem, Salmon,
  Pisseloup, Joffre, Chabot, Retat, Axthelm, Barraclough, Ratcliffe, Bernal,
  Chaumette, Pollini, and Steyn}]{Forshaw2016Removedebris}
Forshaw, J.~L., Aglietti, G.~S., Navarathinam, N., Kadhem, H., Salmon, T.,
  Pisseloup, A., Joffre, E., Chabot, T., Retat, I., Axthelm, R., Barraclough,
  S., Ratcliffe, A., Bernal, C., Chaumette, F., Pollini, A., and Steyn, W.~H.,
  \enquote{{RemoveDEBRIS}: An in-orbit active debris removal demonstration
  mission,} \emph{Acta Astronautica}, Vol. 127, 2016, p. 448–463.
\newblock \doi{10.1016/j.actaastro.2016.06.018}.

\bibitem[{Sharma and D'Amico(2020)}]{Sharma2020TAES}
Sharma, S., and D'Amico, S., \enquote{Neural Network-Based Pose Estimation for
  Noncooperative Spacecraft Rendezvous,} \emph{IEEE Transactions on Aerospace
  and Electronic Systems}, Vol.~56, No.~6, 2020, pp. 4638--4658.
\newblock \doi{10.1109/TAES.2020.2999148}.

\bibitem[{Park et~al.(2019)Park, Sharma, and D'Amico}]{Park2019AAS}
Park, T.~H., Sharma, S., and D'Amico, S., \enquote{Towards Robust
  Learning-Based Pose Estimation of Noncooperative Spacecraft,} \emph{2019
  AAS/AIAA Astrodynamics Specialist Conference, Portland, Maine}, 2019.

\bibitem[{{Pasqualetto Cassinis} et~al.(2021){Pasqualetto Cassinis}, Fonod,
  Gill, Ahrns, and Gil-Fern\'andez}]{PasqualettoCassinis2021Coupled}
{Pasqualetto Cassinis}, L., Fonod, R., Gill, E., Ahrns, I., and
  Gil-Fern\'andez, J., \enquote{Evaluation of tightly- and loosely-coupled
  approaches in CNN-based pose estimation systems for uncooperative
  spacecraft,} \emph{Acta Astronautica}, Vol. 182, 2021, pp. 189--202.
\newblock \doi{10.1016/j.actaastro.2021.01.035}.

\bibitem[{Black et~al.(2021)Black, Shankar, Fonseka, Deutsch, Dhir, and
  Akella}]{Black2021PoseEstimationCygnus}
Black, K., Shankar, S., Fonseka, D., Deutsch, J., Dhir, A., and Akella, M.,
  \enquote{Real-Time, Flight-Ready, Non-Cooperative Spacecraft Pose Estimation
  Using Monocular Imagery,} \emph{31st AAS/AIAA Space Flight Mechanics
  Meeting}, 2021.

\bibitem[{Chen et~al.(2019)Chen, Cao, Bustos, and Chin}]{Chen2019SatellitePE}
Chen, B., Cao, J., Bustos, {\'A}.~P., and Chin, T.-J., \enquote{Satellite Pose
  Estimation with Deep Landmark Regression and Nonlinear Pose Refinement,}
  \emph{2019 IEEE/CVF International Conference on Computer Vision Workshop
  (ICCVW)}, 2019, pp. 2816--2824.
\newblock \doi{10.1109/ICCVW.2019.00343}.

\bibitem[{Proen\c{c}a and Gao(2020)}]{Proenca2019Photorealistic}
Proen\c{c}a, P.~F., and Gao, Y., \enquote{Deep Learning for Spacecraft Pose
  Estimation from Photorealistic Rendering,} \emph{2020 IEEE International
  Conference on Robotics and Automation (ICRA)}, 2020, pp. 6007--6013.
\newblock \doi{10.1109/ICRA40945.2020.9197244}.

\bibitem[{D'Amico et~al.(2013)D'Amico, Bodin, Delpech, and
  Noteborn}]{PRISMA_chapter}
D'Amico, S., Bodin, P., Delpech, M., and Noteborn, R., \enquote{{PRISMA},}
  \emph{Distributed Space Missions for Earth System Monitoring Space Technology
  Library}, Vol.~31, edited by M.~D'Errico, 2013, Chap.~21, pp. 599--637.
\newblock \doi{10.1007/978-1-4614-4541-8\_21}.

\bibitem[{Kisantal et~al.(2020)Kisantal, Sharma, Park, Izzo, M\"{a}rtens, and
  D'Amico}]{Kisantal2020SPEC}
Kisantal, M., Sharma, S., Park, T.~H., Izzo, D., M\"{a}rtens, M., and D'Amico,
  S., \enquote{Satellite Pose Estimation Challenge: Dataset, Competition Design
  and Results,} \emph{IEEE Transactions on Aerospace and Electronic Systems},
  Vol.~56, No.~5, 2020, pp. 4083--4098.
\newblock \doi{10.1109/TAES.2020.2989063}.

\bibitem[{Sharma et~al.(2019)Sharma, Park, and D'Amico}]{Sharma2019SPEEDonSDR}
Sharma, S., Park, T.~H., and D'Amico, S., \enquote{Spacecraft Pose Estimation
  Dataset ({SPEED}),} Stanford Digital Repository., 2019.
\newblock \doi{10.25740/dz692fn7184}, available at:
  \url{https://purl.stanford.edu/dz692fn7184}.

\bibitem[{Pasqualetto~Cassinis et~al.(2020)Pasqualetto~Cassinis, Fonod, Gill,
  Ahrns, and Fernandez}]{PasqualettoCassinis2020CNNEKF}
Pasqualetto~Cassinis, L., Fonod, R., Gill, E., Ahrns, I., and Fernandez, J.~G.,
  \enquote{CNN-Based Pose Estimation System for Close-Proximity Operations
  Around Uncooperative Spacecraft,} \emph{AIAA Scitech 2020 Forum}, 2020.
\newblock \doi{10.2514/6.2020-1457}.

\bibitem[{{Pasqualetto Cassinis} et~al.(2022){Pasqualetto Cassinis}, Menicucci,
  Gill, Ahrns, and Sanchez-Gestido}]{PasqualettoCassinis2022ORGL}
{Pasqualetto Cassinis}, L., Menicucci, A., Gill, E., Ahrns, I., and
  Sanchez-Gestido, M., \enquote{On-ground validation of a CNN-based monocular
  pose estimation system for uncooperative spacecraft: Bridging domain shift in
  rendezvous scenarios,} \emph{Acta Astronautica}, Vol. 196, 2022, pp.
  123--138.
\newblock \doi{10.1016/j.actaastro.2022.04.002}.

\bibitem[{Ben-David et~al.(2010)Ben-David, Blitzer, Crammer, Kulesza, Pereira,
  and Vaughan}]{BenDavid2010LearningFromDiffDomains}
Ben-David, S., Blitzer, J., Crammer, K., Kulesza, A., Pereira, F., and Vaughan,
  J.~W., \enquote{A Theory of Learning from Different Domains,}
  \emph{Mach.~Learn.}, Vol.~79, 2010, p. 151–175.
\newblock \doi{10.1007/s10994-009-5152-4}.

\bibitem[{Peng et~al.(2018)Peng, Usman, Kaushik, Wang, Hoffman, and
  Saenko}]{Peng2017VisDA}
Peng, X., Usman, B., Kaushik, N., Wang, D., Hoffman, J., and Saenko, K.,
  \enquote{VisDA: A Synthetic-to-Real Benchmark for Visual Domain Adaptation,}
  \emph{2018 IEEE/CVF Conference on Computer Vision and Pattern Recognition
  Workshops (CVPRW)}, 2018, pp. 2102--2105.
\newblock \doi{10.1109/CVPRW.2018.00271}.

\bibitem[{Park et~al.(2022)Park, M{\"a}rtens, Lecuyer, Izzo, and
  D'Amico}]{Park2021speedplus}
Park, T.~H., M{\"a}rtens, M., Lecuyer, G., Izzo, D., and D'Amico, S.,
  \enquote{{SPEED+}: Next-Generation Dataset for Spacecraft Pose Estimation
  across Domain Gap,} \emph{2022 IEEE Aerospace Conference (AERO)}, 2022, pp.
  1--15.
\newblock \doi{10.1109/AERO53065.2022.9843439}.

\bibitem[{Park et~al.(2021{\natexlab{a}})Park, M{\"a}rtens, Lecuyer, Izzo, and
  D'Amico}]{Park2021speedplusSDR}
Park, T.~H., M{\"a}rtens, M., Lecuyer, G., Izzo, D., and D'Amico, S.,
  \enquote{Next Generation Spacecraft Pose Estimation Dataset ({SPEED}+),}
  Stanford Digital Repository, 2021{\natexlab{a}}.
\newblock \doi{10.25740/wv398fc4383}, available at
  \url{https://purl.stanford.edu/wv398fc4383}.

\bibitem[{Park et~al.(2023)Park, M{\"a}rtens, Jawaid, Wang, Chen, Chin, Izzo,
  and D'Amico}]{park2023spec2021}
Park, T.~H., M{\"a}rtens, M., Jawaid, M., Wang, Z., Chen, B., Chin, T.-J.,
  Izzo, D., and D'Amico, S., \enquote{Satellite Pose Estimation Competition
  2021: Results and Analyses,} \emph{Acta Astronautica}, Vol. 204, 2023, pp.
  640--665.
\newblock \doi{10.1016/j.actaastro.2023.01.002}.

\bibitem[{Park and D'Amico(2023)}]{park2022spnv2}
Park, T.~H., and D'Amico, S., \enquote{Robust multi-task learning and online
  refinement for spacecraft pose estimation across domain gap,} \emph{Advances
  in Space Research}, 2023.
\newblock \doi{10.1016/j.asr.2023.03.036}.

\bibitem[{Jackson et~al.(2019)Jackson, Atapour-Abarghouei, Bonner, Breckon, and
  Obara}]{Jackson2019ICCV_StyleAug}
Jackson, P.~T., Atapour-Abarghouei, A., Bonner, S., Breckon, T.~P., and Obara,
  B., \enquote{Style Augmentation: Data Augmentation via Style Randomization,}
  \emph{Proceedings of the IEEE Conference on Computer Vision and Pattern
  Recognition Workshops}, 2019, pp. 83--92.

\bibitem[{{Julier} and {Uhlmann}(2004)}]{Julier2004UnscentedFiltering}
{Julier}, S.~J., and {Uhlmann}, J.~K., \enquote{Unscented filtering and
  nonlinear estimation,} \emph{Proceedings of the IEEE}, Vol.~92, No.~3, 2004,
  pp. 401--422.
\newblock \doi{10.1109/JPROC.2003.823141}.

\bibitem[{Tweddle and Saenz-Otero(2015)}]{Tweddle2015MEKF}
Tweddle, B.~E., and Saenz-Otero, A., \enquote{Relative Computer Vision-Based
  Navigation for Small Inspection Spacecraft,} \emph{Journal of Guidance,
  Control, and Dynamics}, Vol.~38, No.~5, 2015, pp. 969--978.
\newblock \doi{10.2514/1.G000687}.

\bibitem[{Sharma and D'Amico(2017)}]{Sharma2017AASGCC}
Sharma, S., and D'Amico, S., \enquote{Reduced-Dynamics Pose Estimation for
  Non-Cooperative Spacecraft Rendezvous using Monocular Vision,} \emph{40th
  Annual AAS Guidance and Control Conference, Breckenridge, Colorado}, 2017.

\bibitem[{Crassidis and Markley(2003)}]{Crassidis2003UnscentedAttitude}
Crassidis, J.~L., and Markley, F.~L., \enquote{Unscented Filtering for
  Spacecraft Attitude Estimation,} \emph{Journal of Guidance, Control, and
  Dynamics}, Vol.~26, No.~4, 2003, pp. 536--542.
\newblock \doi{10.2514/2.5102}.

\bibitem[{Schaub and Junkins(1996)}]{Schaub1995GRP}
Schaub, H., and Junkins, J., \enquote{Stereographic Orientation Parameters for
  Attitude Dynamics: A Generalization of the Rodrigues Parameters,}
  \emph{Journal of the Astronautical Sciences}, Vol.~44, No.~01, 1996, pp.
  1--19.

\bibitem[{Stacey and D'Amico(2021)}]{Stacey2021ASNC}
Stacey, N., and D'Amico, S., \enquote{Adaptive and Dynamically Constrained
  Process Noise Estimation for Orbit Determination,} \emph{IEEE Transactions on
  Aerospace and Electronic Systems}, Vol.~57, No.~5, 2021, pp. 2920--2937.
\newblock \doi{10.1109/TAES.2021.3074205}.

\bibitem[{Myers and Tapley(1976)}]{Myers1976CovarianceMatching}
Myers, K., and Tapley, B., \enquote{Adaptive sequential estimation with unknown
  noise statistics,} \emph{IEEE Transactions on Automatic Control}, Vol.~21,
  No.~4, 1976, pp. 520--523.
\newblock \doi{10.1109/TAC.1976.1101260}.

\bibitem[{Stacey and D'Amico(2022)}]{Stacey2022CovModel}
Stacey, N., and D'Amico, S., \enquote{Analytical process noise covariance
  modeling for absolute and relative orbits,} \emph{Acta Astronautica}, Vol.
  194, 2022, pp. 34--47.
\newblock \doi{10.1016/j.actaastro.2022.01.020}.

\bibitem[{Park and D'Amico(2022)}]{park2022shirt}
Park, T.~H., and D'Amico, S., \enquote{SHIRT: Satellite Hardware-In-the-loop
  Rendezvous Trajectories Dataset,} Stanford Digital Repository, 2022.
\newblock \doi{10.25740/zq716br5462}, available at
  \url{https://purl.stanford.edu/zq716br5462}.

\bibitem[{Park et~al.(2021{\natexlab{b}})Park, Bosse, and
  D'Amico}]{Park2021AAS}
Park, T.~H., Bosse, J., and D'Amico, S., \enquote{Robotic Testbed for
  Rendezvous and Optical Navigation: Multi-Source Calibration and Machine
  Learning Use Cases,} \emph{2021 AAS/AIAA Astrodynamics Specialist Conference,
  Big Sky, Vitrual}, 2021{\natexlab{b}}.

\bibitem[{Tan and Le(2019)}]{Tan2019EfficientNetICML}
Tan, M., and Le, Q., \enquote{{E}fficient{N}et: Rethinking Model Scaling for
  Convolutional Neural Networks,} \emph{Proceedings of the 36th International
  Conference on Machine Learning}, Proceedings of Machine Learning Research,
  Vol.~97, edited by K.~Chaudhuri and R.~Salakhutdinov, PMLR, 2019, pp.
  6105--6114.
\newblock \urlprefix\url{https://proceedings.mlr.press/v97/tan19a.html}.

\bibitem[{Tan et~al.(2020)Tan, Pang, and Le}]{Tan2020EfficientDetCVPR}
Tan, M., Pang, R., and Le, Q.~V., \enquote{{EfficientDet}: Scalable and
  Efficient Object Detection,} \emph{2020 IEEE/CVF Conference on Computer
  Vision and Pattern Recognition (CVPR)}, 2020, pp. 10778--10787.
\newblock \doi{10.1109/CVPR42600.2020.01079}.

\bibitem[{Bukschat and Vetter(2020)}]{Bukschat2020EfficientPose}
Bukschat, Y., and Vetter, M., \enquote{EfficientPose: An efficient, accurate
  and scalable end-to-end 6D multi object pose estimation approach,}
  \emph{CoRR}, Vol. abs/2011.04307, 2020.

\bibitem[{Sharma and D'Amico(2016)}]{Sharma2016CompAssessment}
Sharma, S., and D'Amico, S., \enquote{Comparative assessment of techniques for
  initial pose estimation using monocular vision,} \emph{Acta Astronautica},
  Vol. 123, 2016, pp. 435--445.
\newblock \doi{10.1016/j.actaastro.2015.12.032}.

\bibitem[{Lepetit et~al.(2008)Lepetit, Moreno-Noguer, and
  Fua}]{Lepetit2008EPnP}
Lepetit, V., Moreno-Noguer, F., and Fua, P., \enquote{{EPnP}: An Accurate
  {O(n)} Solution to the {PnP} Problem,} \emph{International Journal of
  Computer Vision}, Vol.~81, No.~2, 2008, p. 155–166.
\newblock \doi{10.1007/s11263-008-0152-6}.

\bibitem[{Wu and He(2018)}]{Wu2018ECCV_GroupNorm}
Wu, Y., and He, K., \enquote{Group Normalization,} \emph{Computer Vision --
  ECCV 2018}, edited by V.~Ferrari, M.~Hebert, C.~Sminchisescu, and Y.~Weiss,
  Springer International Publishing, Cham, 2018, pp. 3--19.
\newblock \doi{10.1007/978-3-030-01261-8_1}.

\bibitem[{Beierle et~al.(2018)Beierle, Norton, Macintosh, and
  D'Amico}]{Beierle2018TwoStageAttitude}
Beierle, C., Norton, A., Macintosh, B., and D'Amico, S., \enquote{{Two-stage
  attitude control for direct imaging of exoplanets with a CubeSat telescope},}
  \emph{Space Telescopes and Instrumentation 2018: Optical, Infrared, and
  Millimeter Wave}, Vol. 10698, International Society for Optics and Photonics,
  SPIE, 2018, pp. 630 -- 644.
\newblock \doi{10.1117/12.2314233}.

\bibitem[{Koenig et~al.(2017)Koenig, Guffanti, and D'Amico}]{Koenig2017STM}
Koenig, A.~W., Guffanti, T., and D'Amico, S., \enquote{New State Transition
  Matrices for Spacecraft Relative Motion in Perturbed Orbits,} \emph{Journal
  of Guidance, Control, and Dynamics}, Vol.~40, No.~7, 2017, pp. 1749--1768.
\newblock \doi{10.2514/1.G002409}.

\bibitem[{Capuano et~al.(2018)Capuano, Kim, Hu, Harvard, and
  Chung}]{Capuano2018Pose}
Capuano, V., Kim, K., Hu, J., Harvard, A., and Chung, S.-J.,
  \enquote{Monocular-based pose determination of uncooperative known and
  unknown space objects,} \emph{69th International Astronautical Congress
  (IAC), Bremen, Germany}, 2018.

\bibitem[{{Markley} and {Crassidis}(2014)}]{Markley2014SADCTextbook}
{Markley}, F.~L., and {Crassidis}, J.~L., \emph{Fundamentals of Spacecraft
  Attitude Determination and Control}, Springer New York, NY, 2014.
\newblock \doi{10.1007/978-1-4939-0802-8}.

\bibitem[{Fraser and Ulrich(2021)}]{Fraser2021AdaptiveKalman}
Fraser, C.~T., and Ulrich, S., \enquote{Adaptive extended Kalman filtering
  strategies for spacecraft formation relative navigation,} \emph{Acta
  Astronautica}, Vol. 178, 2021, pp. 700--721.
\newblock \doi{10.1016/j.actaastro.2020.10.016}.

\bibitem[{Bani~Younes and Mortari(2019)}]{Younes2019AttErrDynamics}
Bani~Younes, A., and Mortari, D., \enquote{Derivation of All Attitude Error
  Governing Equations for Attitude Filtering and Control,} \emph{Sensors},
  Vol.~19, No.~21, 2019.
\newblock \doi{10.3390/s19214682}.

\bibitem[{Markley(2003)}]{Markley2003AttErrRep}
Markley, F.~L., \enquote{Attitude Error Representations for Kalman Filtering,}
  \emph{Journal of Guidance, Control, and Dynamics}, Vol.~26, No.~2, 2003, pp.
  311--317.
\newblock \doi{10.2514/2.5048}.

\bibitem[{D'Amico(2010)}]{Damico2010PHDThesis}
D'Amico, S., \enquote{Autonomous formation flying in low earth orbit,} Ph.D.
  thesis, Technical University of Delft, 2010.

\bibitem[{D'Amico et~al.(2014)D'Amico, Benn, and
  J{\o}rgensen}]{Damico2014IJSSE}
D'Amico, S., Benn, M., and J{\o}rgensen, J.~L., \enquote{Pose estimation of an
  uncooperative spacecraft from actual space imagery,} \emph{International
  Journal of Space Science and Engineering}, Vol.~2, No.~2, 2014, p. 171.
\newblock \doi{10.1504/ijspacese.2014.060600}.

\bibitem[{Dormand and Prince(1980)}]{Dormand1980DOPRI}
Dormand, J., and Prince, P., \enquote{A family of embedded Runge-Kutta
  formulae,} \emph{Journal of Computational and Applied Mathematics}, Vol.~6,
  No.~1, 1980, pp. 19--26.
\newblock \doi{10.1016/0771-050X(80)90013-3}.

\bibitem[{Ries et~al.(2016)Ries, Bettadpur, Eanes, Kang, Ko, McCullough, Nagel,
  Pie, Poole, Richter, Save, and Tapley}]{Ries2016GGM05}
Ries, J.~C., Bettadpur, S.~V., Eanes, R.~J., Kang, Z., Ko, U.-D., McCullough,
  C.~M., Nagel, P., Pie, N., Poole, S.~R., Richter, T., Save, H., and Tapley,
  B.~D., \enquote{The Development and Evaluation of the Global Gravity Model
  {GGM05},} \emph{CSR-16-02, Center for Space Research, The University of Texas
  at Austin}, 2016.

\bibitem[{Picone et~al.(2002)Picone, Hedin, Drob, and
  Aikin}]{Picone2002NRLMSISE00}
Picone, J.~M., Hedin, A.~E., Drob, D.~P., and Aikin, A.~C.,
  \enquote{{NRLMSISE}-00 empirical model of the atmosphere: Statistical
  comparisons and scientific issues,} \emph{Journal of Geophysical Research:
  Space Physics}, Vol. 107, No. A12, 2002, pp. SIA 15--1--SIA 15--16.
\newblock \doi{10.1029/2002JA009430}.

\bibitem[{Gill and Montenbruck(2013)}]{Montenbruck2013SatelliteOrbits}
Gill, E. K.~A., and Montenbruck, O., \emph{Satellite orbits: Models, methods
  and applications}, Springer, 2013.
\newblock \doi{10.1007/978-3-642-58351-3}.

\bibitem[{Wertz(1978)}]{Wertz1978ADCS}
Wertz, J.~R. (ed.), \emph{Spacecraft Attitude Determination and Control},
  Springer Dordrecht, 1978.
\newblock \doi{10.1007/978-94-009-9907-7}.

\bibitem[{Alken et~al.(2020)Alken, Erwan, Beggan, Amit, Aubert, Baerenzung,
  Bondar, Brown, Califf, Chambout, Chulliat, Cox, Finlay, Fournier, Gillet,
  Grayver, Hammer, Holschneider, Huder, and Zhou}]{Alken2020IGRF-13}
Alken, P., Erwan, T., Beggan, C., Amit, H., Aubert, J., Baerenzung, J., Bondar,
  T., Brown, W., Califf, S., Chambout, A., Chulliat, A., Cox, G., Finlay, C.,
  Fournier, A., Gillet, N., Grayver, A., Hammer, M., Holschneider, M., Huder,
  L., and Zhou, B., \enquote{International Geomagnetic Reference Field: the
  thirteenth generation,} \emph{Earth, Planets and Space}, Vol.~73, 2020.
\newblock \doi{10.1186/s40623-020-01288-x}.

\bibitem[{Giralo and D'Amico(2018)}]{Giralo2018GNSSTestbed}
Giralo, V., and D'Amico, S., \enquote{Development of the Stanford GNSS
  Navigation Testbed for Distributed Space Systems,} \emph{Institute of
  Navigation, International Technical Meeting, Reston, Virginia}, 2018.
\newblock \doi{10.33012/2018.15544}.

\bibitem[{Vic(2021)}]{Vicon}
\enquote{Vero: Compact Super Wide Camera by Vicon,}
  \url{https://www.vicon.com/hardware/cameras/vero/}, 2021.
\newblock Accessed April 29, 2021.

\bibitem[{Lig(2016)}]{LightBox}
\enquote{Verification of Light-box Devices for Earth Albedo Simulation,}
  Technical Note, Stanford Space Rendezvous Lab (SLAB), January 2016.

\bibitem[{Sharma et~al.(2018)Sharma, Beierle, and
  D'Amico}]{SharmaBeierle2018_CNN}
Sharma, S., Beierle, C., and D'Amico, S., \enquote{Pose estimation for
  non-cooperative spacecraft rendezvous using convolutional neural networks,}
  \emph{2018 IEEE Aerospace Conference}, 2018, pp. 1--12.
\newblock \doi{10.1109/AERO.2018.8396425}.

\bibitem[{Beierle and D'Amico(2019)}]{Beierle2019}
Beierle, C., and D'Amico, S., \enquote{Variable-Magnification Optical
  Stimulator for Training and Validation of Spaceborne Vision-Based
  Navigation,} \emph{Journal of Spacecraft and Rockets}, Vol.~56, 2019, pp.
  1--13.
\newblock \doi{10.2514/1.A34337}.

\bibitem[{Hauschild and Montenbruck(2021)}]{Hauschild2021GNSSBroadcast}
Hauschild, A., and Montenbruck, O., \enquote{Precise real-time navigation of
  LEO satellites using GNSS broadcast ephemerides,} \emph{NAVIGATION}, Vol.~68,
  No.~2, 2021, pp. 419--432.
\newblock \doi{10.1002/navi.416}.

\bibitem[{Liebe(1995)}]{Liebe1995StarTrackers}
Liebe, C., \enquote{Star trackers for attitude determination,} \emph{IEEE
  Aerospace and Electronic Systems Magazine}, Vol.~10, No.~6, 1995, pp. 10--16.
\newblock \doi{10.1109/62.387971}.

\bibitem[{Christiansen and Nilson(2008)}]{Christiansen2008DockingSM}
Christiansen, S.~S., and Nilson, T., \enquote{Docking System Mechanism Utilized
  on Orbital Express Program,} \emph{Proceedings of the 39th Aerospace
  Mechanisms Symposium, NASA Marshall Space Flight Center, May 7--9}, 2008.

\end{thebibliography}

\end{document}